\documentclass[journal]{IEEEtran}

\usepackage{amsmath,graphicx}
\usepackage{amsfonts}
\usepackage{amsmath,bm}
\usepackage{adjustbox}
\usepackage{algorithmicx}
\usepackage{algpseudocode}
\usepackage[ruled]{algorithm}
\usepackage{graphicx}
\usepackage{caption}
\usepackage{graphicx,times,amsmath} 
\usepackage{caption}
\usepackage{subcaption}
\usepackage[rightcaption]{sidecap}
\usepackage{caption}
\usepackage{multirow}
\usepackage{hhline}
\usepackage{amsmath}
\usepackage{epstopdf}
\usepackage{tabularx}
\usepackage{enumerate}
\usepackage{scrextend}
\usepackage{wrapfig}
\usepackage{caption}
\usepackage{subcaption}
\usepackage{color,soul}

\usepackage[table]{xcolor}

\usepackage{enumitem}

\usepackage{amsfonts}
\usepackage{amsmath}
\usepackage{adjustbox}
\usepackage{algorithmicx}
\usepackage{algpseudocode}
\usepackage[ruled]{algorithm}
\usepackage{graphicx}
\usepackage{caption}
\usepackage{graphicx,times,amsmath} 
\usepackage{makecell}

\usepackage{caption}
\usepackage{subcaption}
\usepackage[rightcaption]{sidecap}
\usepackage{caption}
\usepackage{multirow}
\usepackage{hhline}
\usepackage{amsmath}
\usepackage{epstopdf}
\usepackage{tabularx}
\usepackage{algorithmicx}
\usepackage{algpseudocode}
\usepackage[ruled]{algorithm}

\usepackage{epstopdf}
\usepackage{tabularx}
\usepackage{subcaption}

\usepackage{tabularx,booktabs}
\newcolumntype{C}{>{\centering\arraybackslash}X} 
\usepackage{lipsum}

\makeatletter
\def\old@comma{,}
\catcode`\,=13
\def,{%
  \ifmmode%
    \old@comma\discretionary{}{}{}%
  \else%
    \old@comma%
  \fi%
}
\makeatother

\graphicspath{ {images/} }

\usepackage{alphalph}

\title{EDropout: Energy-Based Dropout and Pruning of Deep Neural Networks}

\author{Hojjat~Salehinejad, \textit{Member, IEEE}, and Shahrokh~Valaee, \textit{Fellow, IEEE}
        
\thanks{H. Salehinejad and S. Valaee are with the Department of Electrical \& Computer Engineering, University of Toronto, Toronto, Canada e-mail: hojjat.salehinejad@mail.utoronto.ca, valaee@ece.utoronto.ca.

This article has been accepted for inclusion in a future issue of the IEEE TRANSACTIONS ON NEURAL NETWORKS AND LEARNING SYSTEMS.  

\copyright 2021 IEEE. Personal use of this material is permitted. Permission
 from IEEE must be obtained for all other uses, in any current or future
media, including reprinting/republishing this material for advertising or
 promotional purposes, creating new collective works, for resale or
 redistribution to servers or lists, or reuse of any copyrighted
 component of this work in other works. 
 
 Digital Object Identifier 10.1109/TNNLS.2021.3069970}
\thanks{}
\thanks{}}

\begin{document}
\newcommand*{\img}{%
  \includegraphics[
    width=\linewidth,
    height=20pt,
    keepaspectratio=false,
  ]{example-image-a}%
}

\maketitle

\begin{abstract}
Dropout is a well-known regularization method by sampling a sub-network from a larger deep neural network and training different sub-networks on different subsets of the data. Inspired by the dropout concept, we propose EDropout as an energy-based framework for pruning neural networks in classification tasks. In this approach, a set of binary pruning state vectors (population) represents a set of corresponding sub-networks from an arbitrary  original neural network. An energy loss function assigns a scalar energy loss value to each pruning state. The energy-based model stochastically evolves the population to find states with lower energy loss. The best pruning state is then selected and applied to the original network. Similar to dropout, the kept weights are updated using backpropagation in a probabilistic model. The energy-based model again searches for better pruning states and the cycle continuous. This procedure is in fact a switching between the energy model, which manages the pruning states, and the probabilistic model, which updates the kept weights, in each iteration. 
The population can dynamically converge to a pruning state. This can be interpreted as dropout leading to pruning the network. From an implementation perspective, unlike most of the pruning methods, EDropout can prune neural networks without manually modifying the network architecture code. We have evaluated the proposed method on different flavours of ResNets, AlexNet, $l_{1}$ pruning, ThinNet, ChannelNet, and SqueezeNet on the Kuzushiji, Fashion, CIFAR-10, CIFAR-100, Flowers, and ImageNet datasets, and compared the pruning rate and classification performance of the models. The networks trained with EDropout on average achieved a pruning rate of more than $50\%$ of the trainable parameters with approximately $<5\%$ and $<1\%$ drop of Top-1 and Top-5 classification accuracy, respectively.  
\end{abstract}


\section{Introduction}
\label{sec:intro}

Deep neural networks (DNNs) have different learning capacities based on the number of trainable parameters for classification tasks. Depending on the complexity of the dataset (e.g. size of dataset and number of classes), selecting a network which maximizes the generalization performance often involves trial and error. A DNN with larger capacity often has better performance, but may suffer from a longer training time, lack of generalization, and redundancy in trained parameters. DNNs are computationally intensive, require large memory, and consume extensive energy~\cite{han2015deep}. Hence, smaller networks are more desirable for applications such as edge computing and embedded systems. Pruning DNNs is one of the major approaches for reducing the number of trainable parameters while maintaining the classification performance. Most of the proposed methods for deploying DNN on edge devices can be divided into pruning existing networks or small networks by design. Our focus in this paper is on methods that can prune a given DNN.

Dropout \cite{srivastava2014dropout}, which was originally proposed as a regularization technique to train DNNs, reduces complexity of the network by randomly dropping a subset of the parameters in each training phase but fully utilizing the parameters (full network) for inference. Despite other regularization methods such as $l_{1}$ and $l_{2}$ norms~\cite{hastie2015statistical}, which modify the loss function, the dropout-based methods modify the network structure, which is equivalent to training different smaller neural networks. A variety of methods has been proposed for \textit{smarter} dropout such as standout~\cite{ba2013adaptive}, variational dropout~\cite{kingma2015variational}, and adversarial dropout~\cite{park2018adversarial}. Dropout is mainly proposed for dense layers while it is much less used in convolution layers. While dropout~\cite{srivastava2014dropout} sets a subset of activation values to zero, \textit{DropConnect}~\cite{wan2013regularization} zeroes a randomly selected subset of weights in a fully connected network. A survey on dropout methods is provided in~\cite{labach2019survey}.

Pruning is different from dropout, in which the trainable parameters are permanently dropped, and is mainly used to remove redundant weights while preserving the classification accuracy. Removing inactive and redundant weights from a network can lead to a smaller network which has a better generalization performance and a faster inference speed~\cite{lecun1990optimal}. In general, pruning algorithms have three stages which are training, pruning, and fine-tuning~\cite{liu2018rethinking}. One of the early attempts was to use second derivative information to minimize a cost function that reduces network complexity by removing excess number of trainable parameters and further training the remaining of the network to increase inference accuracy~\cite{lecun1990optimal}. Soft weight-sharing~\cite{nowlan1992simplifying} is another approach by clustering weights into subgroups with similar weight values. It adds a penalty term to the cost function of the network where the distribution of weight values is modeled as a mixture of multiple Gaussians and the means and variance of clusters are adapted during the training of network~\cite{nowlan1992simplifying}. This work has been further enhanced in~\cite{ullrich2017soft}. \textit{Deep Compression} has three stages that are pruning, quantization, and Huffman coding, which targets reducing the storage requirement of a DNN without affecting its accuracy~\cite{han2015deep}. This method works by pruning all connections with weights below a threshold followed by retraining the sparsified network.

The concept of energy has also been used for pruning DNNs, where we have proposed an Ising energy-based model for dropout and pruning of hidden units in multi-layer perceptron (MLP) neural networks~\cite{salehinejad2019ising}. In this approach, the network is modeled as a graph where each node is a hidden unit and the edge between nodes represents a measure of activation of each unit. We used a high-speed hardware to search for the best sub-network using Markov Chain Monte-Carlo (MCMC)~\cite{matsubara2017ising} method. Later, we scaled up this method for pruning large-scale MLPs using a node grouping technique~\cite{salehinejad2019isingGlobal} and pruning  DCNNs~\cite{salehinejad2021pruning}.

\begin{figure}[t]
\centering
\captionsetup{font=footnotesize}
\includegraphics[width=0.4\textwidth]{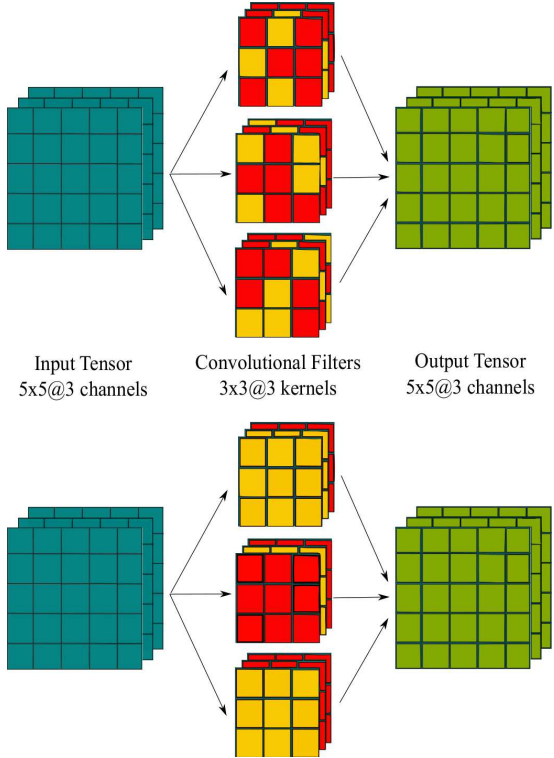}  
\caption{Top: Unstructured pruning; Bottom: Structured pruning. Red elements are pruned.}
\label{fig:struc_unstruc_pruning}
\vspace{-4mm}
\end{figure}

In this paper, we introduce the concept of using energy-based models (EBMs)~\cite{lecun2006tutorial} for pruning of convolutional and dense layers. This method behaves as a hybrid technique, where similar to random dropout, it trains different sub-networks of the original DNN at a time, leading to a pruned network. The trained network is in fact a subset of the original network, which has a competitive performance to the network at full capacity but with a smaller number of parameters. We introduce utilizing a population of pruning candidate states, where each state vector in the population represents a set of active trainable weights from the network. Each vector has a corresponding energy loss value as a measure of dependency between sub-network parameters. The objective is to search for state vectors with lower energy loss. Indeed, for a DNN with $|\mathbf{\Theta}|$ number of trainable parameters, this is a combinatorial optimization problem which requires $2^{|\mathbf{\Theta}|}$ times evaluation of the energy loss function for each batch of data. Since this procedure is NP-hard, we also propose a stochastic parallel binary technique based on the differential evolution (DE) method~\cite{price2013differential}, \cite{salehinejad2017micro} to search for the sparse state of the network. 
In this paper, our focus is to study and evaluate EDropout in pruning DNNs and not as a regularization method. Our results show on average more than $50\%$ pruning of trainable network parameters while maintaining classification performance --- the Top-1 and Top-5 classification accuracy drop on average $<5\%$ and $<1\%$, respectively --- on several image classification tasks with various number of classes and available training samples on the ResNets, AlexNet, $l_{1}$ pruning, ThinNet, ChannelNet, and SqueezeNet.

\begin{figure}[t]
\centering
\captionsetup{font=footnotesize}
\includegraphics[width=0.45\textwidth]{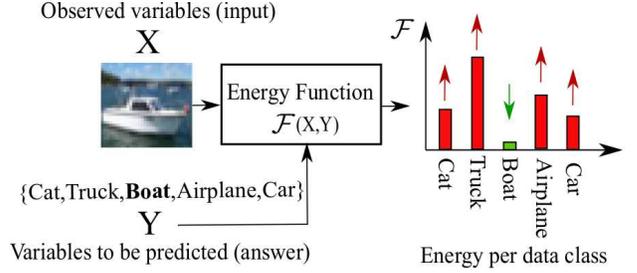}  
\caption{Energy-based model (EBM),~\cite{lecun2006tutorial}.}
\label{fig:energy_model_diagram}
\vspace{-4mm}
\end{figure}

\section{Background}
\subsection{Pruning Methods}
Pruning methods in neural networks can be divided into unstructured and structured methods. Figure~\ref{fig:struc_unstruc_pruning} shows an illustration of these two approaches.

\subsubsection{Unstructured Pruning}
Unstructured pruning removes any subset of weights without following a specific geometry like the entire kernel, filter, or channel~\cite{anwar2017structured}. This kind of pruning results in a sub-network with geometrically sparse weights, which is difficult to implement in practice as a smaller network due to predefined tensor-based operations on graphical processing units (GPUs). It also requires overhead computation to address the location of kept weights. Therefore, it is difficult to get computational advantage from this approach. However, due to granular level of sparsity, this approach generally has lower drop of accuracy compared with the structured methods and the original model. 

\subsubsection{Structured Pruning}
Structured pruning generally follows some constraints and defined structure in pruning a network. Typically pruning happens at channel, kernel, and intra-kernel levels. All the incoming and outgoing weights to/from a feature map are pruned in a channel level pruning which can result in an intensive reduction of network weights. In kernel level pruning, a subset of kernels is entirely pruned. 
Compared with the unstructured pruning, the structured pruning has very little  computational cost overhead and is easier to implement and scale on GPUs. The advantage is faster implementation and inference.
A group of state-of-the-art structured pruning algorithms is examined in~\cite{liu2018rethinking}, which concludes that fine-tuning a pruned model only gives comparable or worse performance than training the model with randomly initialized weights.

 \begin{figure*}[!t]
\centering
\captionsetup{font=small}
\includegraphics[width=0.8\textwidth]{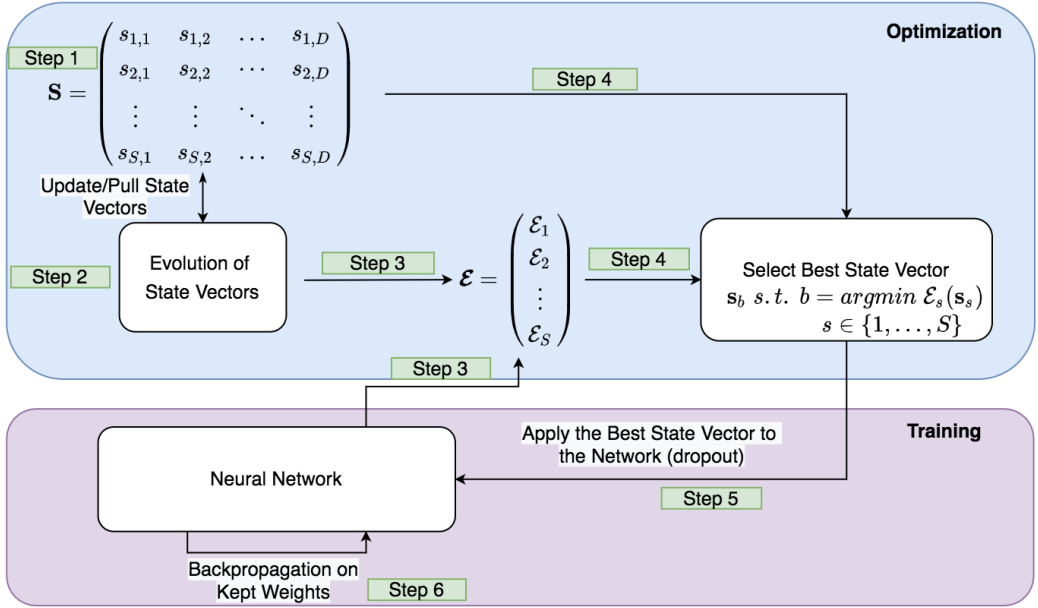}  
\caption{Different steps of the proposed combined optimization and training framework for pruning neural networks. Step 1: Population initialization; Step 2: Evolution of the population; Step 3: Compute energy loss value for each candidate state vector; Step 4: Select the state vector with the lowest energy loss value in the population; Step 5: Apply the state vector with lowest energy loss value (the best state vector) to the network; Step 6: Perform backpropagation (on the kept weights only); Step 7: Continue to step 2. $\mathbf{S}$: population of candidate pruning state vectors; $\mathbf{s}_{b}$: best pruning state vector; $\mathcal{E}_{s}$: energy loss of state vector $s$.}
\label{fig:cnn_example_EPruning}
\vspace{-4mm}
\end{figure*}

\subsection{Energy-based Models}
In general, solving a classification problem using DNNs with $C$ data classes $\mathcal{Y}=\{y_{1},...,y_{C}\}$ is defined as using a parametric function to map the input $X\in\mathcal{X}$ to $C$ real-valued numbers $\epsilon=\{\varepsilon_{1},\varepsilon_{2},...,\varepsilon_{C}\}$ (a.k.a. logits). This function is in fact a stack of convolutional and dense layers with non-linear activation functions. The output is then passed to a classifier, such as \textit{Softmax} function, which is a member of the \textit{Gibbs distributions}, to parameterize a categorical distribution in form of a probability distribution over the data classes~\cite{grathwohl2019your}, defined as $\{p(y_{1}),...,p(y_{C})\}$ where for simplicity we define $p_{c}=p(y_{c})\:\forall \: c\in\{1,...,C\}$. 
The loss is then calculated based on cross-entropy with respect to the correct answer $Y$. 

The Gibbs distribution (a.k.a the Boltzmann distribution), which is a very general family of probability distributions, is defined as 
\begin{equation}
p(Y|X)=\frac{e^{-\beta \mathcal{F}(Y,X)}}{Z(\beta)},
\label{eq:gibbs}
\end{equation}
where $Z(\beta)=\sum\limits_{y_{c}\in\mathcal{Y}}e^{-\beta \mathcal{F}(y_{c},X)}$ is the partition function, $\beta>0$ is the inverse temperature parameter~\cite{lecun2006tutorial}, and $\mathcal{F}(\cdot)$ is referred to as \textit{Hamiltonian} or the \textit{energy function}. The Gibbs distribution naturally permits annealing, by changing the temperature during the training procedure~\cite{labach2020framework}. $\beta$ is discussed in~\cite{hinton2015distilling} as the \textit{distillation} where a higher value for $1/\beta$ produces a softer
probability distribution over classes.

 In general, some probabilistic models can be considered as a special type of EBMs. An EBM assigns a \textit{scalar energy loss} as a measure of compatibility to a configuration of parameters in neural networks~\cite{lecun2006tutorial}, as demonstrated in Figure~\ref{fig:energy_model_diagram}. One of the advantages of this approach is avoiding normalization, which can be interpreted as an alternative to probabilistic estimation~\cite{lecun2006tutorial}.
 Calculating the exact probability in~(\ref{eq:gibbs}) needs computing the partition function over all the data classes $C$. However, for large $C$, such as in language models with more than $100,000$ classes, this causes a bottleneck~\cite{barber2016dealing}. In addition, in order to follow the probability axioms, specifically $\sum_{c=1}^{C}p_{c}=1$, the Softmax transfer function generates prior probability values that are close to zero for each class. Some methods such as annealed importance sampling~\cite{neal2001annealed} and negative sampling~\cite{mikolov2013distributed} have been proposed to facilitate computing of the partition function, which is out of the scope of this paper.

\subsection{Differential Evolution}
Population-based optimization algorithms are popular methods for solving combinatorial high-dimensional optimization problems. The advantage of such methods is using a population of candidate solutions to search the problem landscape to minimize/maximize the objective function. DE~\cite{price2013differential} is one of the popular and well-known population based algorithms which has been used in different applications such as localization~\cite{salehinejad20143d} and opposition-based learning~\cite{salehinejad2014type}. The search procedure starts with $S$ initial candidate solutions (individuals) $\mathbf{S}^{S\times D}$ with dimension $D$ and based on the scaled difference between two selected individuals 
$\mathbf{s}_{i},\mathbf{s}_{j \neq i}\in\mathbf{S}$
improves the candidate solutions in each generation toward a feasible solution~\cite{salehinejad2017micro}. 
The standard version of DE for \textit{continuous problems} has three major operations: mutation, crossover, and selection.

\textit{Mutation:} For an individual $\mathbf{s}_{i}\in\mathbf{S}$, three different individuals $\mathbf{s}_{i_{1}}$, $\mathbf{s}_{i_{2}}$, and $\mathbf{s}_{i_{3}}$ are selected from $\mathbf{S}$. The mutant vector of $\mathbf{s}_{i}$ is then calculated as
\begin{equation}
\mathbf{v}_{i}=\mathbf{s}_{i_{1}}+F(\mathbf{s}_{i_{2}}-\mathbf{s}_{i_{3}}),
\end{equation}  
where the mutation factor $F\in(0.1,1.5)$ is a real constant that controls the amplification of the added differential variation of $\textbf{s}_{i_{2}}-\textbf{s}_{i_{3}}$. The exploration of DE increases by selecting higher values for $F$.
 
\textit{Crossover:} The crossover operation increases diversity of the population by shuffling the mutant and parent vectors as 
\begin{equation}
u_{i,d}= \left\{ \begin{array}{ll} v_{i,d}, & r'_{d}\le C_{r}\: \text{or}\: d_{rand}=d\\
s_{i,d}, & \textrm{otherwise}
\end{array},\right.
\end{equation}
for all $d\in\{1,...,D\}$ and $i\in\{1,...,S\}$, where $C_{r}\in[0,1]$ is the crossover rate, $r'_{d}\in[0,1]$, and $d_{rand}$ is a random integer from the interval $[1,D]$. 

\textit{Selection:} The $\mathbf{u}_{i}$ and $\mathbf{s}_{i}$ vectors are evaluated for each ${\mathbf{s}_{i}\in\mathbf{S}}$ and compared with respect to their fitness value. The one with better fitness (lower value for minimization problem) is selected for the next generation. 

The above procedure generally continues for a predefined number of generations or until some convergence criteria are satisfied.

\begin{figure}[!t]
\centering
\captionsetup{font=footnotesize}
\includegraphics[width=0.4\textwidth]{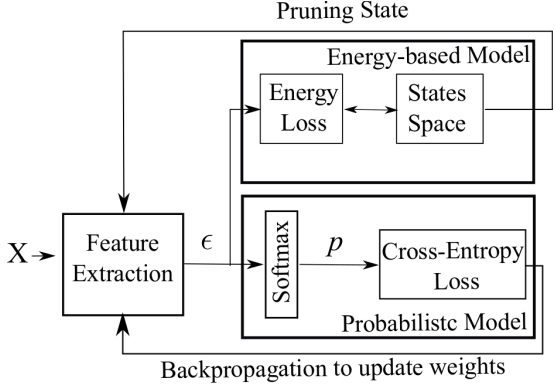}  
\caption{Switching between the energy-based model (EBM) and the probabilistic model. The EBM searches for the pruning state and the probabilistic models searches for the weights. Both models are aware of target class $Y$ during training. In inference, the best pruning state is applied and the EBM is removed.}
\label{fig:energy_model_block}
\vspace{-4mm}
\end{figure}

\section{Energy-based Dropout and Pruning}

Figure~\ref{fig:cnn_example_EPruning} shows an overview of the proposed framework. It has two major phases which are optimization and training. The optimization phase manages searching for a state vector for dropout and ultimately pruning a neural network. The training phase trains a sub-network of the given neural network defined by the state vector suggested by the optimization phase. These two phases are conducted interchangeably in each iteration until an \textit{early state convergence} condition (discussed in Subsection~\ref{sec:earlystateconvergence}) is met. Then, the optimization phase is terminated and the training phased is focused on fine-tuning the sub-network, represented by the last suggested state vector. Details of these steps are discussed next.


\subsection{Pruning States}
\label{sec:states}
A given neural network has a set of trainable parameters $\mathbf{\Theta}$, which is generally a combination of parameters in convolutional and dense layers. In convolutional layers, we are interested in dropping weight filters (including bias terms) and in dense layers, the hidden units (including the bias terms and all incoming/outgoing connections). A feature map in a convolutional layer is the output of convolving a weight filter with the incoming activation values. Therefore, dropping a filter can also be interpreted as dropping a feature map or an input channel to the next layer. 

Let us define a set of $S$ binary candidate pruning state vectors as the population $\mathbf{S}^{S\times D}$. Each vector $\mathbf{s}_{i}\in\mathbf{S}$ of length $D$ represents the state of trainable filters in a convolution layer and hidden units in the dense layers, referred to as a \textit{unit} hereafter for convenience. If $s_{i,d}=0$ unit $d$ is dropped (inactive) and if $s_{i,d}=1$ it participates (active) during training and inference. 
Each state vector $\mathbf{s}_{i}$ corresponds to a sub-network of the original network with the corresponding energy function values $\mathcal{F}_{i}\in\{\mathcal{F}_{1},...,\mathcal{F}_{S}\}$, where generally we can have $2^{D}$ possible energy functions.

 \begin{figure*}[t]
\centering
\captionsetup{font=small}
\includegraphics[width=0.7\textwidth]{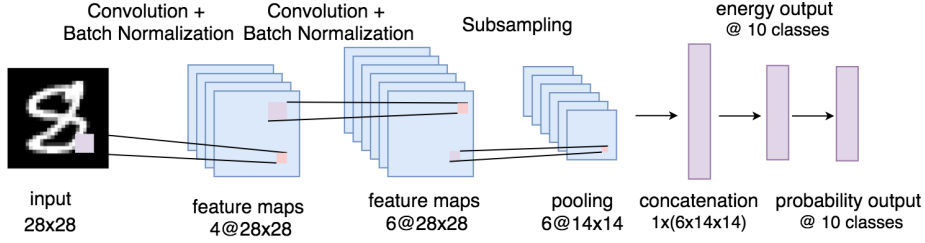}  
\caption{A simple convolutional neural network to visualize the energy landscape in Figure~\ref{fig:energyLossLandscape} with $2^{10}$ possible state vectors, where each state corresponds to a convolution filter.}
\label{fig:simple_network}
\vspace{-5mm}
\end{figure*}

\begin{figure}[!t]
\centering
\centering
\captionsetup{font=footnotesize}
\centering
\includegraphics[width=0.5\textwidth]{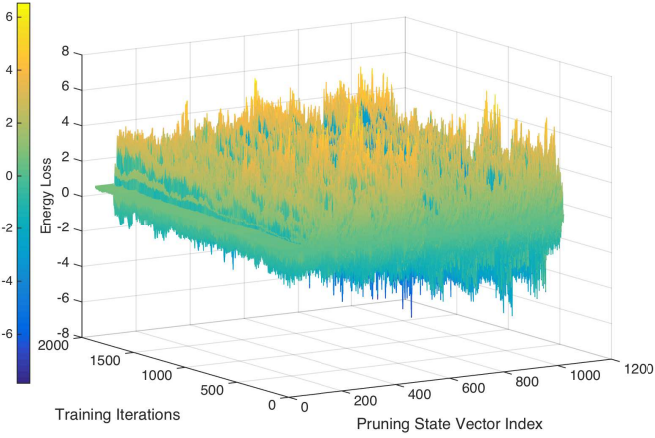}
\caption{Energy loss landscape for all possible pruning state vectors ($2^{10}$) during training iterations of the neural network in Figure~\ref{fig:simple_network}.}
\label{fig:energyLossLandscape}
\vspace{-4mm}
\end{figure}


\subsection{Energy Model}
The Gibbs distribution gives considerable flexibility in defining pruning methods~\cite{labach2020framework}. For example, every Markov random field (MRF) corresponds to a Gibbs distribution, and vice versa.
Gibbs distribution is a probability measure to define the state of a system (such as a DNN) based on the given \textit{energy} of the system. Softmax function which is the normalized exponential function generally used in training DNN for classification problems has the similar form as the Boltzmann distribution~\cite{murphy2012machine}. An energy function is a measure of compatibility, which represents the dependencies of a subset of the network variables as a scalar energy, based on the definition of EBMs in~\cite{lecun2006tutorial}. 

By choosing $\beta=1$ in the Gibbs distribution~(\ref{eq:gibbs}), the energy function corresponding to using a Softmax layer in training neural networks is given by
\begin{equation}
\mathcal{F}(Y,X) = -\epsilon,    
\end{equation}
where $\epsilon=(\varepsilon_{1},...,\varepsilon_{C})$.
The Softmax function in training neural networks is defined as
\begin{equation}
p(y_{c}|X)=\frac{e^{\varepsilon_{c}}}{\sum_{\varepsilon_{c}\in\epsilon}e^{\varepsilon_{c}}}.
\end{equation}
Since Softmax is a special case of the Gibbs distribution, by setting $\beta=1$~\cite{murphy2012machine}, and using the definition in~(\ref{eq:gibbs}) we have
\begin{equation}
p_(y_{c}|X) = \frac{e^{-\mathcal{F}(y_c,X)}}{\sum_{y_c\in\mathcal{Y}}e^{-\mathcal{F}(y_c,X)}},
\label{eq:gibbsAsenergy}
\end{equation}
where by comparing~(\ref{eq:gibbs}) and~(\ref{eq:gibbsAsenergy}) we can interpret ${\mathcal{F}(y_{c},X)=-\varepsilon_{c}}$ and in a generalized form ${\mathcal{F}(Y,X)=-\epsilon}$ for all $\varepsilon_{c}\in\epsilon$. 

In EBMs, training a model refers to finding an energy function $\mathcal{F}(Y,X,\Theta)$ that minimizes the energy loss of the model by searching the weights space~\cite{lecun2006tutorial}. However, we define pruning as the searching for a binary state vector $\mathbf{s}$ that prunes the network while minimizes the energy loss for fixed $Y,X$, and $\Theta$ in each iteration, defined as $\mathcal{F}(Y,X,\Theta,\mathbf{s})$. We are using the cross-entropy loss and backpropagation to search for the weights. Hence, the search for weights is conducted using a probabilistic model when the pruning state vector is fixed and the search for pruning state is conducted using an EBM when the weights are fixed in each iteration, as illustrated in Figure~\ref{fig:energy_model_block}. Based on this concept, the candidate state vectors discussed in Section~\ref{sec:states} define different energy functions corresponding to different sub-networks of the original network. Evolution of the candidate state vectors in an energy minimization scheme helps to find a subset of the original network which corresponds to an energy function with low energy value. 


We define the following energy loss function to measure the quality of an energy function for $(X,Y)$ with the target output $y_{c}$ as
\begin{equation}
\begin{split}
\mathcal{E}&=\mathcal{L}(Y,\mathcal{F})\\
&=\mathcal{F}(y_{c},X)-
min\{\mathcal{F}(y_{c'},X): \; y_{c'}\in\mathcal{Y},c'\neq c\},
\end{split}
\label{eq:energy_loss_function}
\end{equation}
where it can be extended for a batch of data.
The \textit{energy loss} function is intuitively designed to assign a low loss value to $\mathcal{F}_{s}$ which has the lowest energy with respect to the target data class $c$ and higher energy with respect to the other data classes and vice versa~\cite{lecun2006tutorial}. The corresponding energy loss values for the state vectors in Figure~\ref{fig:cnn_example_EPruning} are calculated in Step 3 as $\mathcal{E}=(\mathcal{E}_{1},...,\mathcal{E}_{S})$. 

For sake of visualization, Figure~\ref{fig:simple_network} shows a simple neural network for an image classification task, which has $4$ filters in the first convolution layer and $6$ filters in the second convolution layer. For simplicity, we only focus on pruning the convolution filters. Hence, the length of a state vector $\mathbf{s}$ is $D=10$ such that $s_{d}\in\{0,1\}$. Figure~\ref{fig:energyLossLandscape} shows the complexity of the energy loss landscape of this network for all possible state vectors (i.e. $2^{D}=2^{10}$) over the training iterations, which shows finding a state vector with minimum energy loss value is a very complex NP-hard combinatorial problem with many possible local minima. In the next subsection, we discuss our proposed optimization approach to search for the best state vector on the energy landscape. 


\begin{algorithm}[t]
\small
\begin{algorithmic} 
\State Set t = 0 // Optimization counter
\State Initiate the neural network with trainable weights $\Theta$
\State Set $\mathbf{S}^{(0)}\sim Bernoulli(P=0.5)$ // States initialization
\State Set $\Delta\mathbf{s}\neq 0$ \& $\Delta\mathbf{s}_{T}$
\For{ $ \mathit{i_{epoch}} = 1 \rightarrow \mathit{N_{epoch}}$} // epoch counter
\For{ $ \mathit{i_{batch}} = 1 \rightarrow \mathit{N_{batch}}$} // batch counter
\State t = t+1
\If{$\Delta\mathbf{s}\neq 0$ or $i_{epoch}\leq \Delta\mathbf{s}_{T}$}
\If{$i_{epoch}=1 \; \& \; i_{batch}=1$}
\State Compute energy loss of $\mathbf{S}^{(0)}$ as $\mathcal{E}^{(0)}$ using~(\ref{eq:energy_loss_function})
\EndIf

\For {$i=1\rightarrow S$} // States counter
\State Generate mutually different $i_{1},i_{2},i_{3}\in \{1,...,S\}$
\For {$d=1\rightarrow D$} // State dimension counter
\State Generate a random number $r_{d}\in[0,1]$
\State Compute mutation vector $v_{i,d}$ using (\ref{eq:mutation})
\EndFor
\State Select candidate state $\tilde{s}^{(t)}_{i}$ using (\ref{eq:crossover})
\EndFor
\State Compute energy loss of $\tilde{\mathbf{S}}^{(t)}$ as $\tilde{\mathcal{E}}^{(t)}$ using~(\ref{eq:energy_loss_function})
\State Select $\mathbf{S}^{(t)}$ and corresponding energy $\mathcal{E}^{(t)}$ using (\ref{eq:selection})
\State Select the state with the lowest energy from $\mathbf{S}^{(t)}$ as $\mathbf{s}^{(t)}_{b}$
\Else
\State $\mathbf{s}^{(t)}_{b}=\mathbf{s}^{(t-1)}_{b}$
\EndIf
\State Temporarily drop weights of the network based on the best state $\mathbf{s}^{(t)}_{b}$
\State Compute loss of the sparsified network
\State Perform backpropagation to update $\Theta$
\EndFor
\State Update $\Delta\mathbf{s}$ for early state convergence using (\ref{eq:stateconvergence})
\EndFor
\end{algorithmic}
\small
  \caption{EDropout}
  \label{alg:EPruning}
\end{algorithm}

\subsection{Training and Optimization}
Training and optimization in EDropout have the following major steps which are initialization, energy loss computation, energy loss optimization, early state convergence check, and training the sub-network in each iteration using backpropagation, as presented in Algorithm~\ref{alg:EPruning}.

\subsubsection{Initialization}
At the beginning of training ($t=0$), we initialize the candidate pruning states $\mathbf{S}^{(0)}\in \mathbb{Z}_{2}^{S\times D}$, where $s_{i,d}^{(0)}\sim Bernoulli(P=0.5)$ for $i\in\{1,...,S\}$ and $d\in\{1,...,D\}$.

\subsubsection{Computing Energy Loss}
For each candidate state $\mathbf{s}_{i}^{(t)}\in\mathbf{S}^{(t)}$ in iteration $t$, the energy loss value is calculated using~(\ref{eq:energy_loss_function}) as $\mathcal{E}^{(t)}_{i}$.

\subsubsection{Energy Loss Optimization}
Searching for the pruning state which can minimize the energy loss value is an NP-hard combinatorial problem. Various methods such as 
MCMC and simulated annealing (SA) can be used to search for low energy states. We propose using a binary version of DE (BDE)~\cite{price2013differential} to minimize the energy loss function. This method has the advantage of searching the optimization landscape in parallel and sharing the search experience among candidate states. The other advantage of this approach is flexibility of designing the energy function with constraints.

\begin{figure*}[!t]
\centering
\centering
\captionsetup{font=footnotesize}

\begin{subfigure}[t]{0.5\textwidth}
\captionsetup{font=footnotesize}
\centering
\includegraphics[width=1\textwidth]{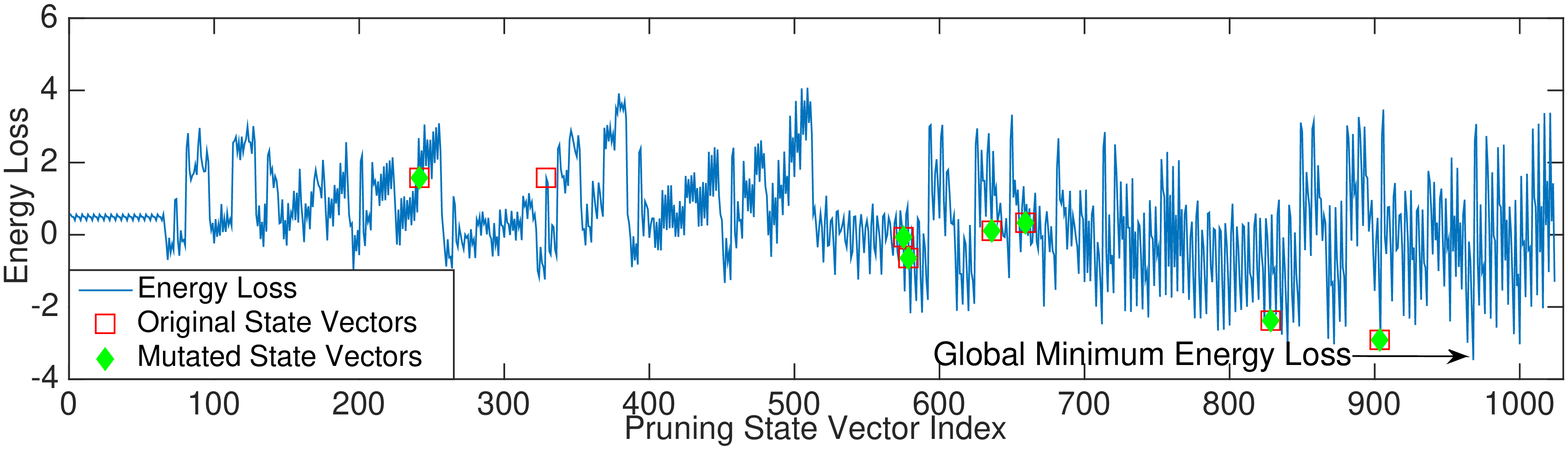}  
\caption{$F=0.1$}
\label{fig:}
\end{subfigure}%
\begin{subfigure}[t]{0.5\textwidth}
\captionsetup{font=footnotesize}
\centering
\includegraphics[width=1\textwidth]{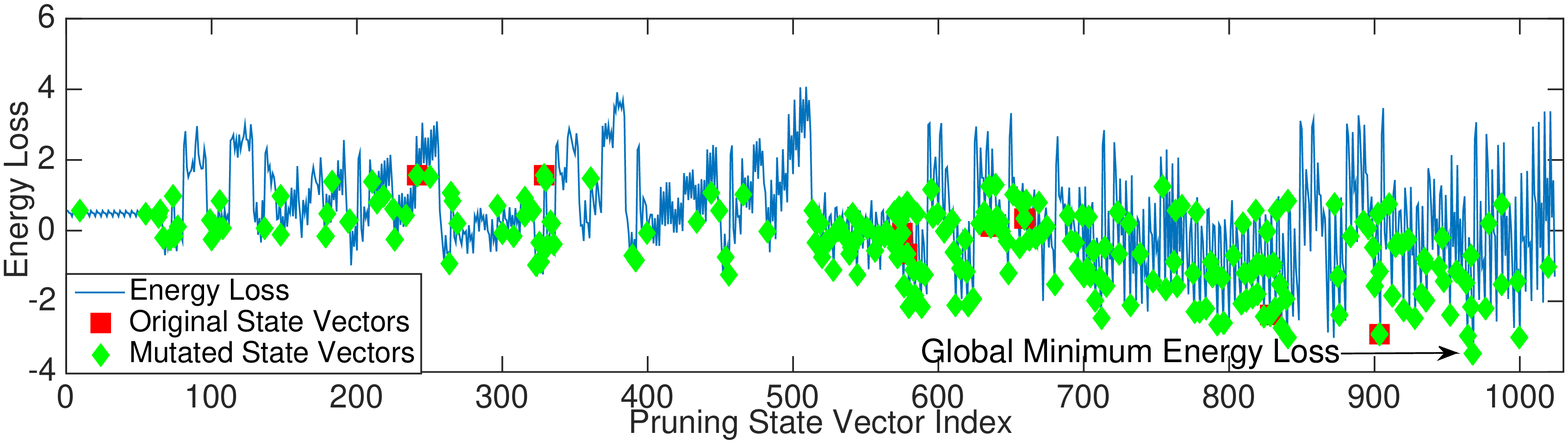}  
\caption{$F=0.1$}
\label{fig:}
\end{subfigure}%

\begin{subfigure}[t]{0.5\textwidth}
\captionsetup{font=footnotesize}
\centering
\includegraphics[width=1\textwidth]{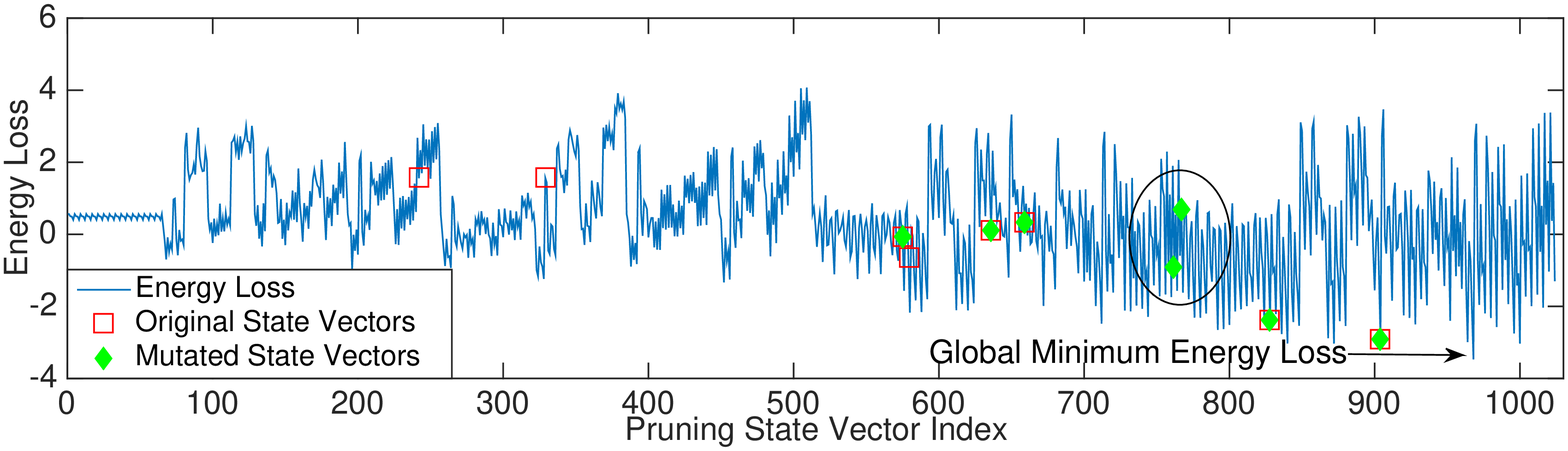}  
\caption{$F=0.9$}
\label{fig:}
\end{subfigure}%
\begin{subfigure}[t]{0.5\textwidth}
\captionsetup{font=footnotesize}
\centering
\includegraphics[width=1\textwidth]{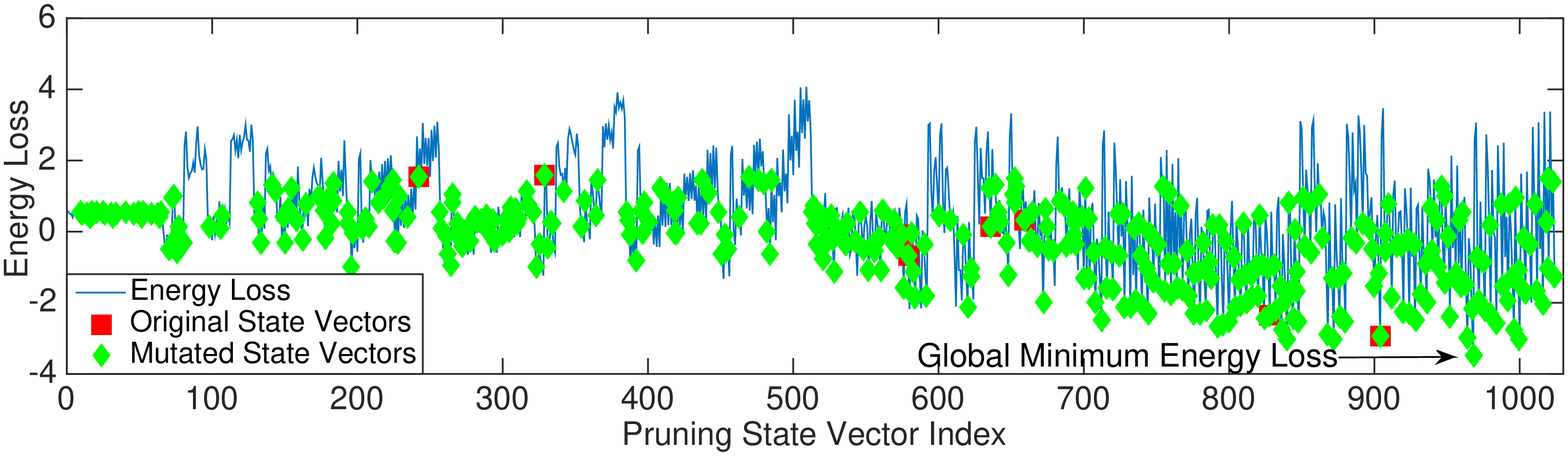}  
\caption{$F=0.9$}
\label{fig:}
\end{subfigure}%

\begin{subfigure}[t]{0.5\textwidth}
\captionsetup{font=footnotesize}
\centering
\includegraphics[width=1\textwidth]{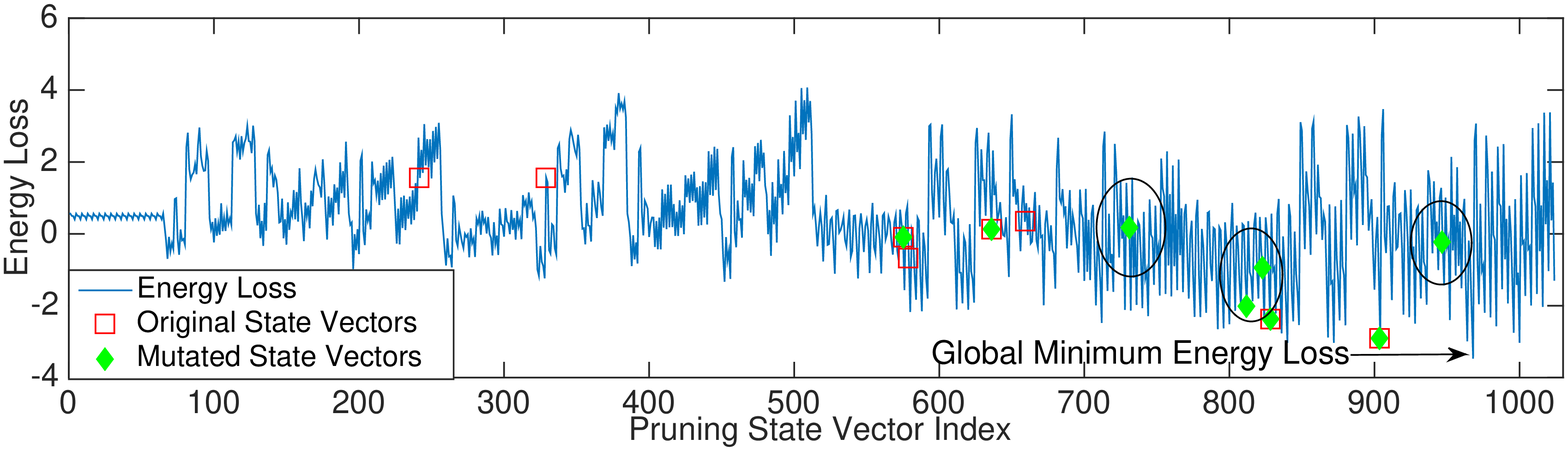}  
\caption{$F\in[0,1]$}
\label{fig:}
\end{subfigure}%
\begin{subfigure}[t]{0.5\textwidth}
\captionsetup{font=footnotesize}
\centering
\includegraphics[width=1\textwidth]{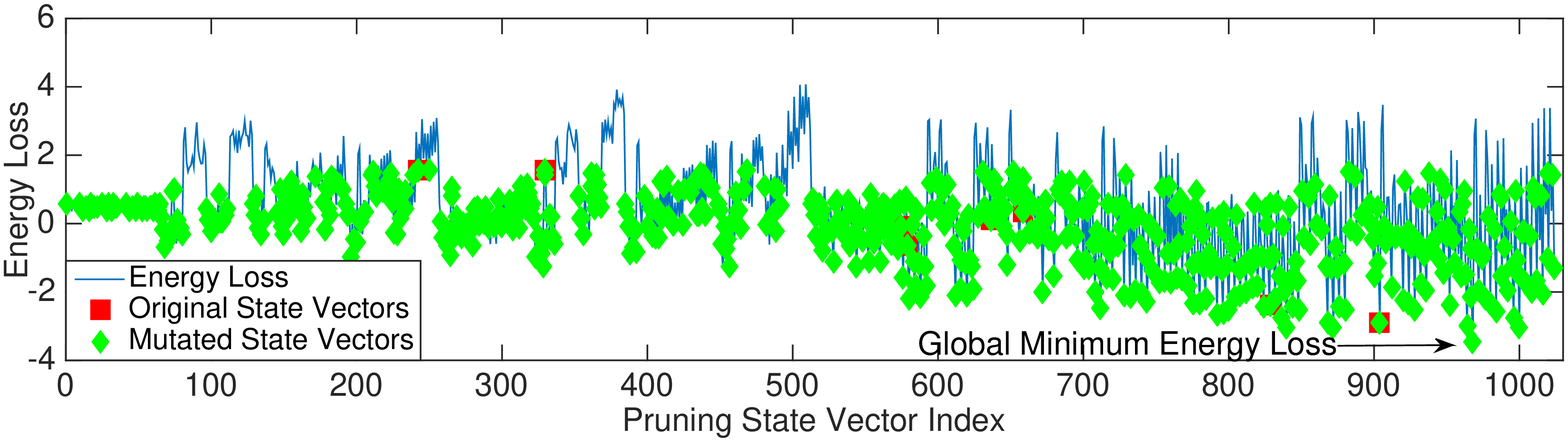}  
\caption{$F\in[0,1]$}
\label{fig:}
\end{subfigure}%
\caption{Left: Visualization of mutated state vectors diversity in a single run. Right: Monte-Carlo simulation of mutated state vectors diversity for 1,000 runs.}
\vspace{-3mm}
\label{fig:visualization_F}
\end{figure*}

The optimization step has three phases which are mutation, crossover, and selection. Given the population of states $\mathbf{S}^{(t-1)}$, a mutation vector is defined for each candidate state ${\mathbf{s}_{i}^{(t-1)}\in\mathbf{S}^{(t-1)}}$ as 
\begin{equation}
v_{i,d}=\begin{cases}
               1-s_{i_{1},d}^{(t-1)}, \:\:\:\:$if$\:\:\:s_{i_{2},d}^{(t-1)}\neq s_{i_{3},d} ^{(t-1)}\;$\&$\; r_{d}<F\\
               s_{i_{1},d}^{(t-1)}, \:\:\:\:\:\:\:\:\:\:\:$ otherwise $
            \end{cases},
\label{eq:mutation}
\end{equation}
for all $d\in\{1,..,D\}$, where $i_{1},i_{2},i_{3}\in \{1,...,S\}$ are mutually different, $F$ is the mutation factor~\cite{salehinejad2017micro}, and $r_{d}\in[0,1]$ is a random number. The next step is to crossover the mutation vectors to generate new candidate state vectors as
\begin{equation}
\tilde{s}^{(t)}_{i,d}=\begin{cases}
               v_{i,d} \:\:\:\:\:\:\:\:\:\:\:\:$if$\:\:\: r'_{d}\in[0,1] \leq C_{r}\\
               s_{i,d}^{(t-1)} \:\:\:\:\:\:\:\:\:\:\:$ otherwise $
            \end{cases},
\label{eq:crossover}
\end{equation}
where $C_{r}$ is the crossover coefficient~\cite{salehinejad2017micro}. The parameters $C_{r}$ and $F$ control exploration and exploitation of the population on the optimization landscape. Each generated state $ \tilde{\mathbf{s}}_{i}^{(t)}$ is then compared with its corresponding parent with respect to its energy loss value $\tilde{\mathcal{E}}^{(t)}_{i}$ as
\begin{equation}
\mathbf{s}_{i}^{(t)}=\begin{cases}
               \tilde{\mathbf{s}}_{i}^{(t)} \:\:\:\:\:\:\:\:\:\:$if$\:\:\:    \tilde{\mathcal{E}}^{(t)}_{i}\leq \mathcal{E}^{(t-1)}_{i} \\
              \mathbf{s}_{i}^{(t-1)} \:\:\:\:\:$ otherwise $
            \end{cases}\:\forall\:i\in\{1,...,S\}.
\label{eq:selection}
\end{equation}
The state with minimum energy loss $\mathcal{E}_{b}^{(t)}=min\{\mathcal{E}_{1}^{(t)},...,\mathcal{E}_{S}^{(t)}\}$ is selected as the best state $\mathbf{s}_{b}$, which represents the sub-network for next training batch. This optimization strategy is simple and feasible to implement in parallel for a large $S$. A predefined number of active states can be defined in the optimizer to enforce a specific dropout/pruning rate.

To better understand the importance of mutation factor, we select the energy loss landscape at iteration 1200 from Figure~\ref{fig:energyLossLandscape}, where the state vector with the lowest energy loss is ${(1,1,1,1,0,0,0,1,1,1)}$ at index $968$ with an energy loss value of $-3.4715$. Then $S=8$ random state vectors are initialized, as illustrated by red markers in Figures~\ref{fig:visualization_F}(a), \ref{fig:visualization_F}(c), and \ref{fig:visualization_F}(e)
for $F=0.1$, $F=0.9$, and $F\in[0,1]$, respectively, and $C_{r}=0.9$. Then, the state vectors are mutated only once, their energy loss is compared with their parent, and the state vector with lower energy loss is denoted by green markers. The plots show $F=0.9$ results in more diversity of mutated state vectors than $F=0.1$. We need to note that some mutated individuals may overlap, as observed for $F=0.1$ and $F=0.9$.
If we use a random $F\in[0,1]$, the diversity is further enhanced compared to when using a constant $F$, and mutated states are more likely to have a lower energy loss value than their parent. In another visualization, we perform a Monte-Carlo simulation of the above experiment with 1,000 independent runs. Comparing  Figures~\ref{fig:visualization_F}(b), \ref{fig:visualization_F}(d), and \ref{fig:visualization_F}(f) show that $F\in[0,1]$ has more exploration capability. In fact, for $F=0.1$, $F=0.9$, and $F\in[0,1]$, about $23.05\%$, $36.91\%$, and $60.35\%$ of the possible states are visited, respectively. 

\begin{figure}[!t]
\centering
\centering
\captionsetup{font=footnotesize}
\begin{subfigure}[t]{0.49\textwidth}
\captionsetup{font=footnotesize}
\centering
\includegraphics[width=1\textwidth]{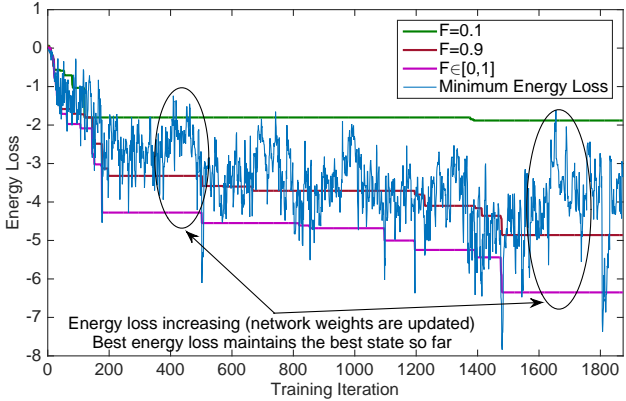}
\caption{Convergence of best mutated state vector for different mutation factors with ${C_{r}=0.5}$.}
\label{fig:}
\end{subfigure}%

\begin{subfigure}[t]{0.49\textwidth}
\captionsetup{font=footnotesize}
\centering
\includegraphics[width=1\textwidth]{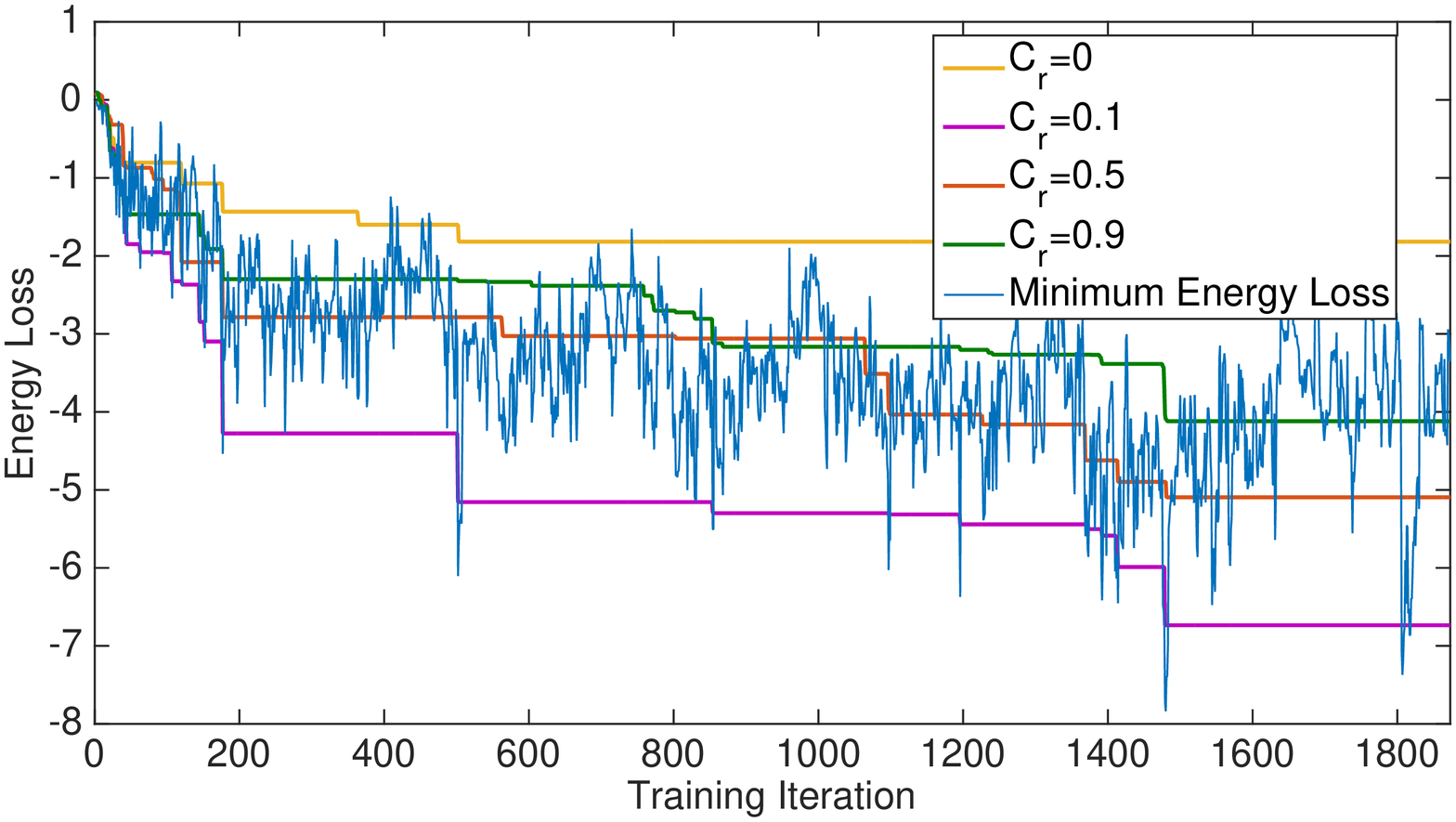}  
\caption{Convergence of best mutated state vector for different crossover rates with ${F\in[0,1]}$.}
\label{fig:}
\end{subfigure}%
\caption{Convergence of best mutated state vector for different mutation factors and crossover rates.}
\vspace{-5mm}
\label{fig:visualization_Cr}
\end{figure}

We also look at the energy loss landscape from the training iterations perspective. As Figure~\ref{fig:visualization_Cr}(a) shows, using the random $F\in[0,1]$ results in a better exploration of the optimization landscape, and ultimately, lower energy loss over the training iterations. It is interesting to observe that although the minimum energy loss value of a single input fluctuates during training (due to backpropagation on all training samples), the energy loss of selected state vectors remains the lowest possible for each iteration and enforces pruning while decreasing the overall energy loss.

The last visualization is about the importance of the crossover rate, as illustrated in Figure~\ref{fig:visualization_Cr}(b) for $F\in[0,1]$. We can observe that the combination of the mutation factor and the crossover rate can control the exploration and exploitation of the search. Too much diversity ($F\in[0,1]$ and $C_{r}=0.9$) 
results in stagnation while a balance of these two parameters ($F\in[0,1]$ and $C_{r}=0.1$), can result in finding the states with lower energy loss.

\subsubsection{Early State Convergence}
\label{sec:earlystateconvergence}
The population-based optimizers are global optimization methods which generally converge to a locally optimal
solution. These algorithms suffer from premature convergence and stagnation problems~\cite{lampinen2000stagnation}. The former generally occurs when the population (candidate state vectors) has converged to local optima, has lost its diversity, or has no improvement in finding better solutions. The latter happens mainly when the population remains diverse during training~\cite{lampinen2000stagnation}.

The optimization process can run for every iterations of the neural network training. However, after a number of iterations, depending on the capacity of the neural network and the complexity of the dataset, all the states in $\mathbf{S}^{(t)}$ may converge to a state $\mathbf{s}_{b}\in\mathbf{S}^{(t)}$. We call this the \textit{early state convergence} phase and define it as 
\begin{equation} 
\Delta\mathbf{s} =
\mathcal{E}_{b}^{(t)} -
\frac{1}{S}\sum\limits_{j=1}^{S}\mathcal{E}_{j}^{(t)},
\label{eq:stateconvergence}
\end{equation}
where $\mathcal{E}_{b}^{(t)}$ is the energy loss of $\mathbf{s}_{b}$.
So, if $\Delta\mathbf{s}=0$ we can call for an \textit{early state convergence} and continue training by fine-tuning the sub-network identified by the state vector $\mathbf{s}_{b}$. In addition, a stagnation threshold $\Delta\mathbf{s}_{T}$ is implemented where if $\Delta\mathbf{s}\neq 0$ after $\Delta\mathbf{s}_{T}$ number of training epochs, it stops the energy loss optimizer.
These mechanisms are implemented to balance exploration and exploitation of the optimizer and address potential stagnation and premature convergence scenarios during training, as analyzed in Section~\ref{sec:exp_earlystateconvergence}.

Most pruning and compression models first prune and train the network and then fine-tune it. The convergence to the best state $\mathbf{s}_{b}$ in \textit{EDropout} breaks the training procedure of the neural network into two phases. The first phase is when various subsets of the neural network, with potential overlap of units, are trained, which occurs before the convergence, and the second phase is when only a subset of network, chosen by $\mathbf{s}_{b}$, is fine-tuned. The first phase is equivalent to training a network with \textit{dropout}, which may act as a regularization method, and the second phase acts as \textit{fine-tuning} where the pruned units/filters are not trained anymore and are practically eliminated from the network. This leads to a \textit{sparsified} neural network. Training the sparsified network is similar to a typical training of DNNs with backpropagation.

\subsubsection{Training}
Once the best candidate state is selected in each iteration, the state vector is applied to the network to compute the training loss. The backpropagation is then performed for the active units.

\begin{figure*}[!t]
\centering
\captionsetup{font=footnotesize}
\begin{subfigure}[t]{0.185\textwidth}
\centering
\captionsetup{font=footnotesize}
\includegraphics[width=1\textwidth]{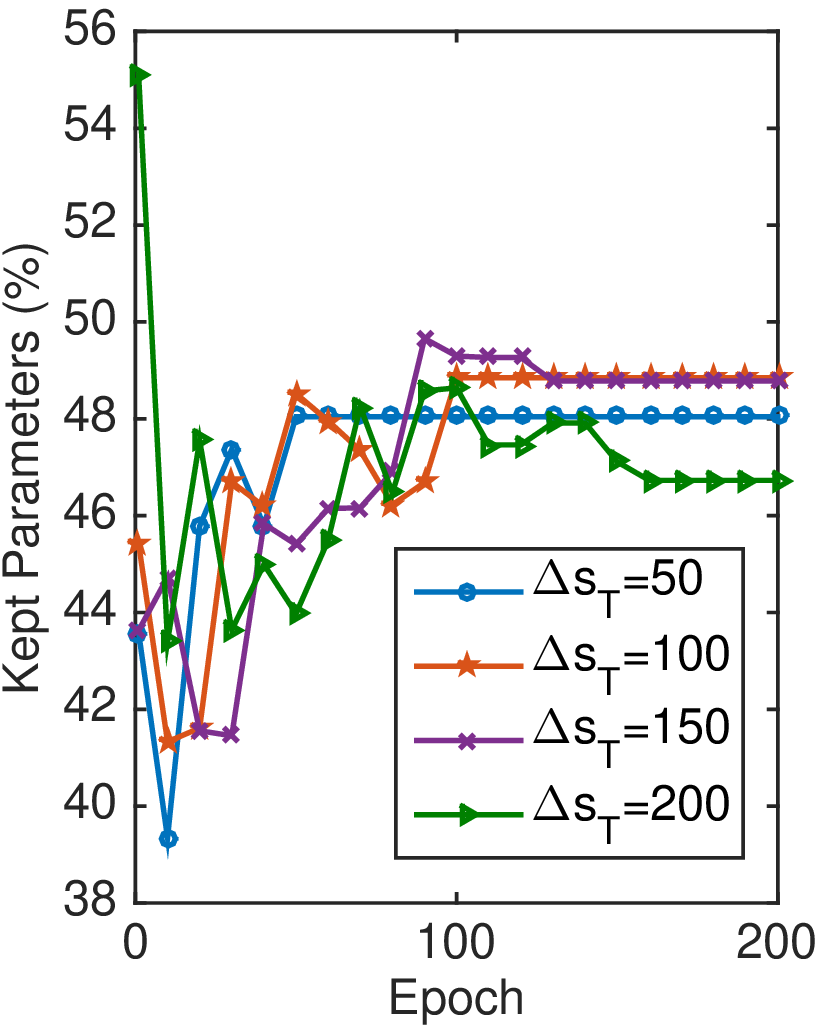}  
\caption{Kept parameters}
\end{subfigure}%
~        
\begin{subfigure}[t]{0.185\textwidth}
\captionsetup{font=footnotesize}
\centering
\includegraphics[width=1\textwidth]{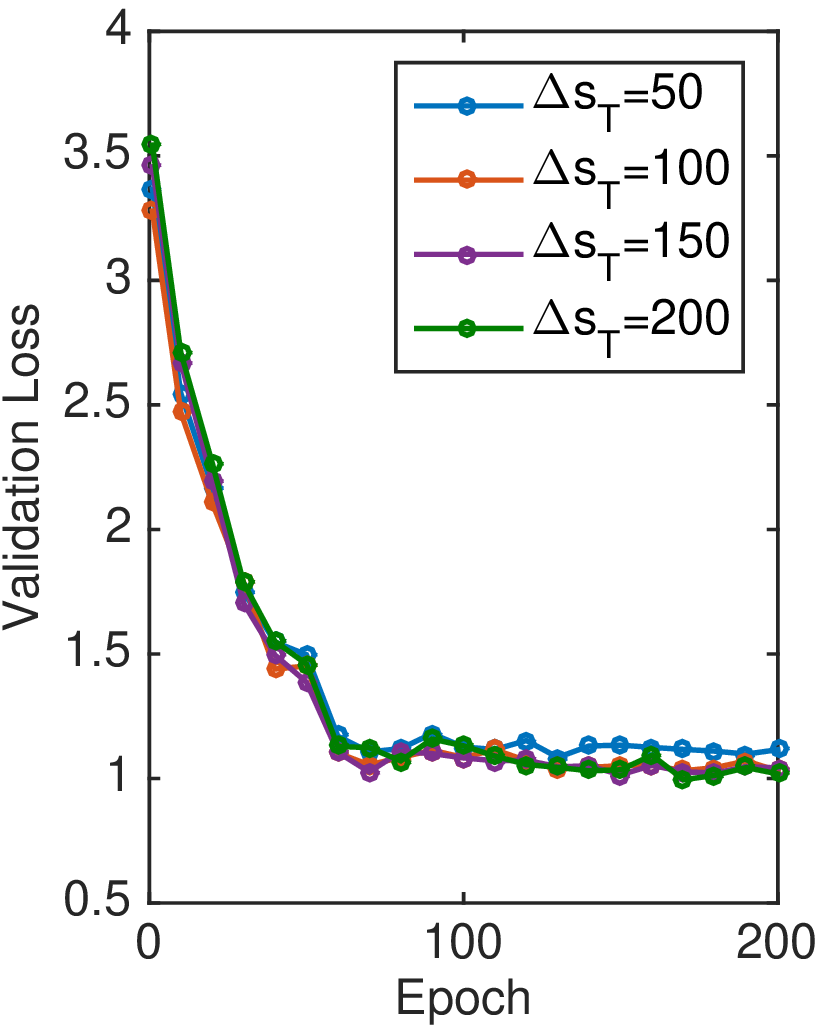}
\caption{Validation Loss}
\end{subfigure}%
~     
\begin{subfigure}[t]{0.185\textwidth}
\captionsetup{font=footnotesize}
\centering
\includegraphics[width=1\textwidth]{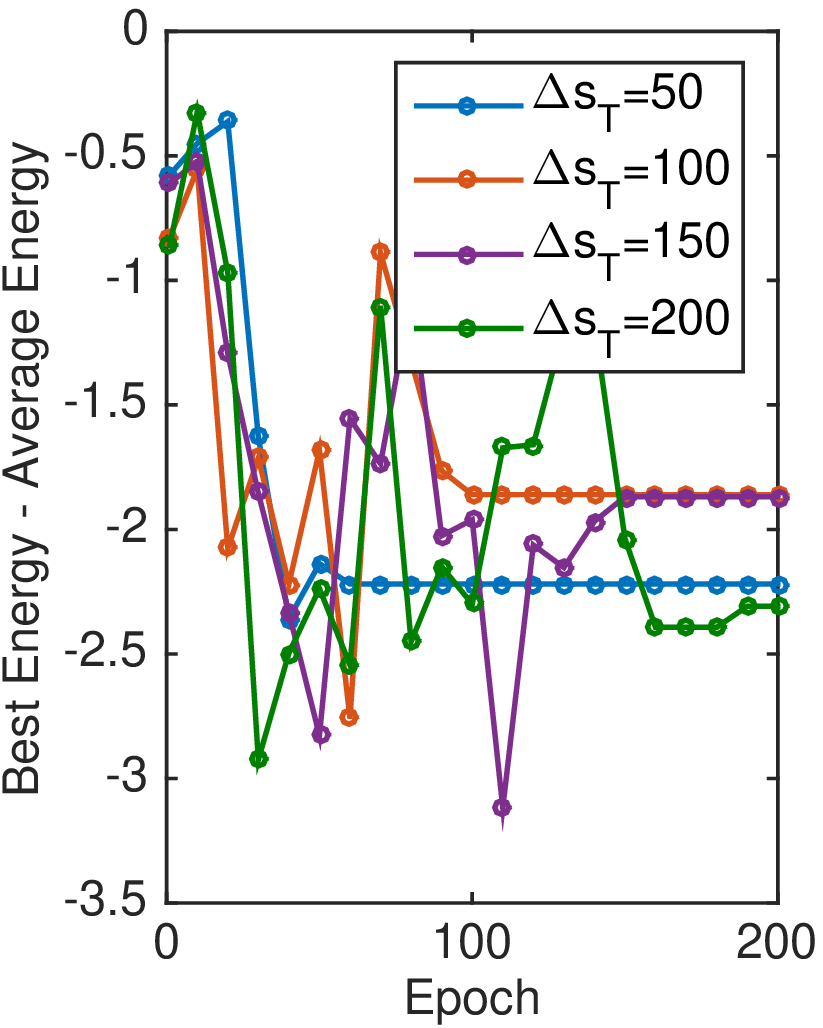}  
\caption{Energy loss difference}
\end{subfigure}%
~
\begin{subfigure}[t]{0.185\textwidth}
\captionsetup{font=footnotesize}
\centering
\includegraphics[width=1\textwidth]{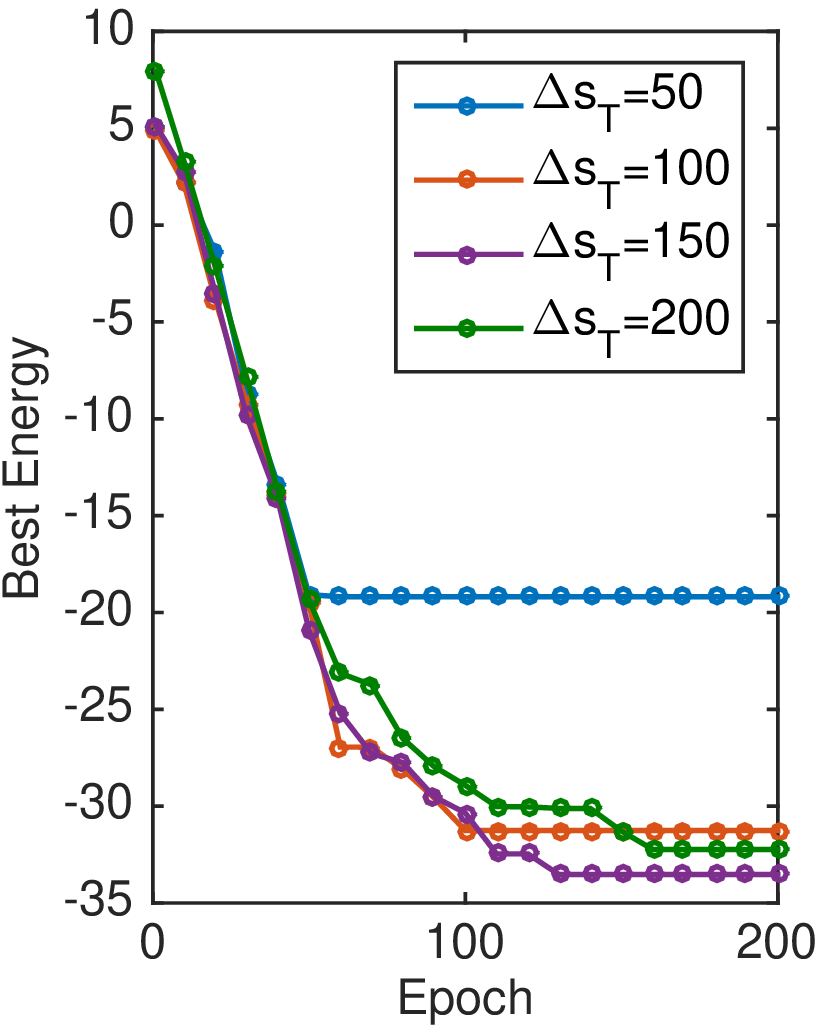}
\caption{Best energy loss}
\end{subfigure}   
~     
\begin{subfigure}[t]{0.185\textwidth}
\captionsetup{font=footnotesize}
\centering
\includegraphics[width=1\textwidth]{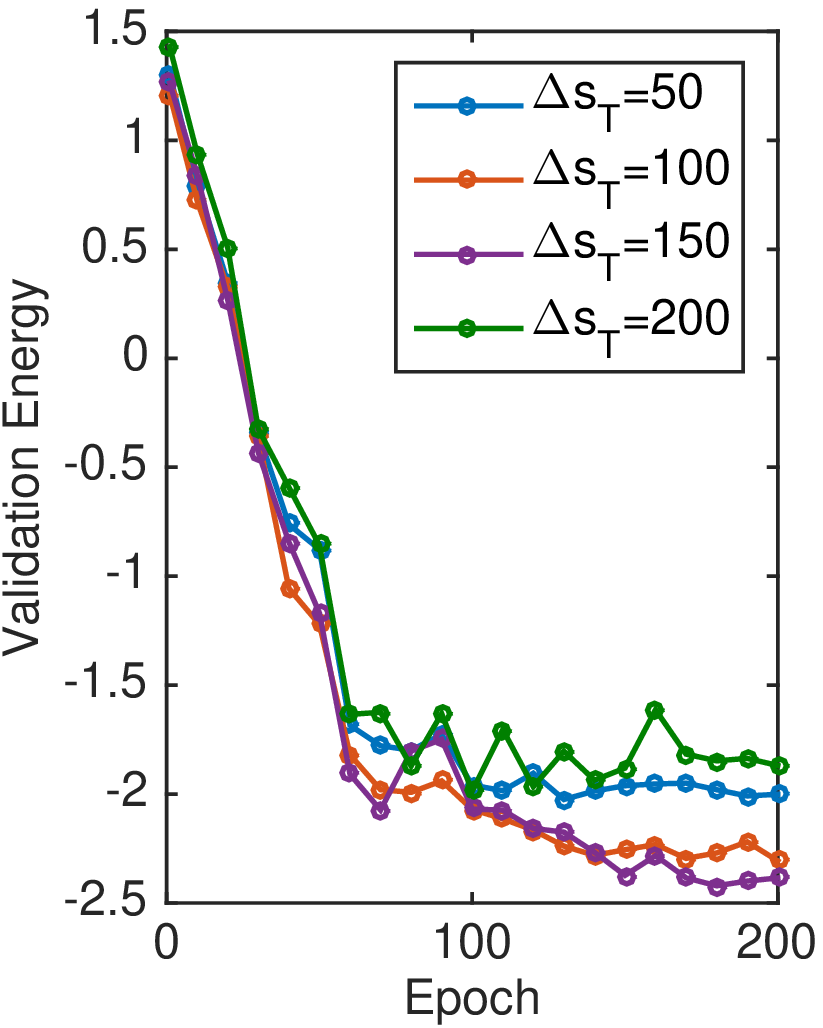}
\caption{Validation energy loss}
\end{subfigure}   
\caption{Convergence of \textbf{ResNet-18} on \textbf{Flowers} validation dataset with $S=8$ and initialization probability of $P=0.5$ over 200 epochs.}
\label{fig:threshold_earlystate_resenet18_flowers}
\end{figure*}

\begin{figure*}
\centering
\begin{minipage}{.49\textwidth}
\centering
\captionsetup{font=footnotesize}
\begin{subfigure}[t]{0.38\textwidth}
\centering
\captionsetup{font=footnotesize}
\includegraphics[width=1\textwidth]{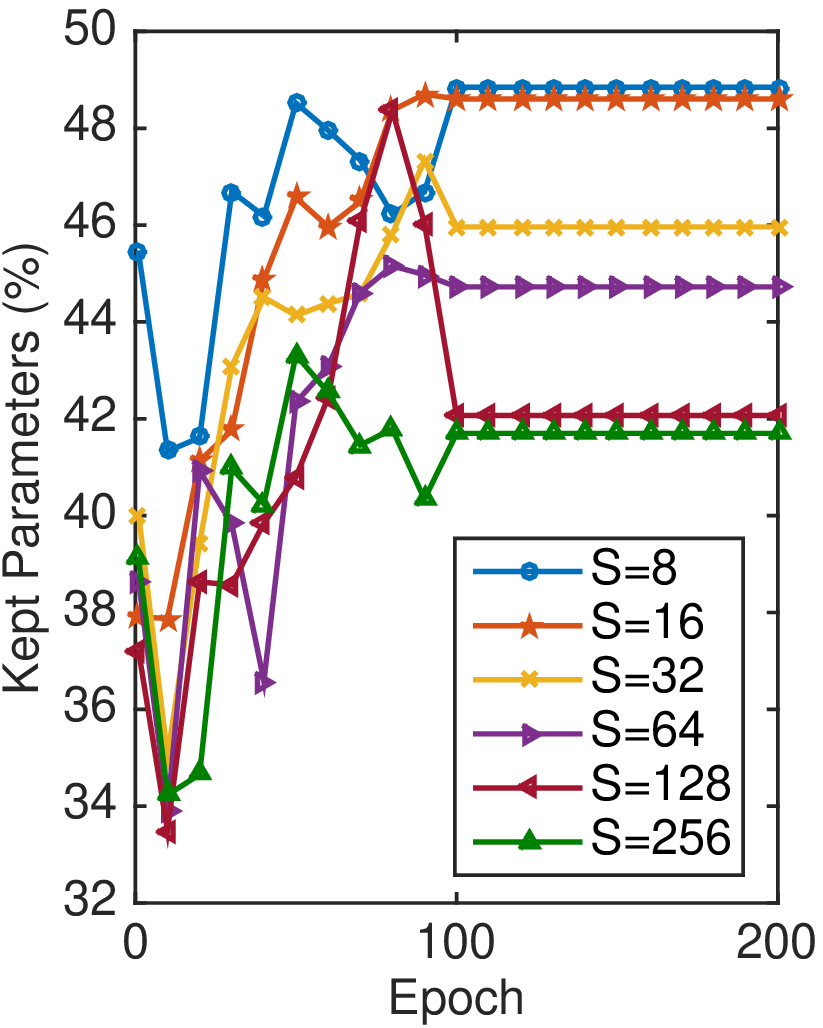}  
\caption{Rate of kept parameters}
\end{subfigure}%
~~~
\begin{subfigure}[t]{0.38\textwidth}
\captionsetup{font=footnotesize}
\centering
\includegraphics[width=1\textwidth]{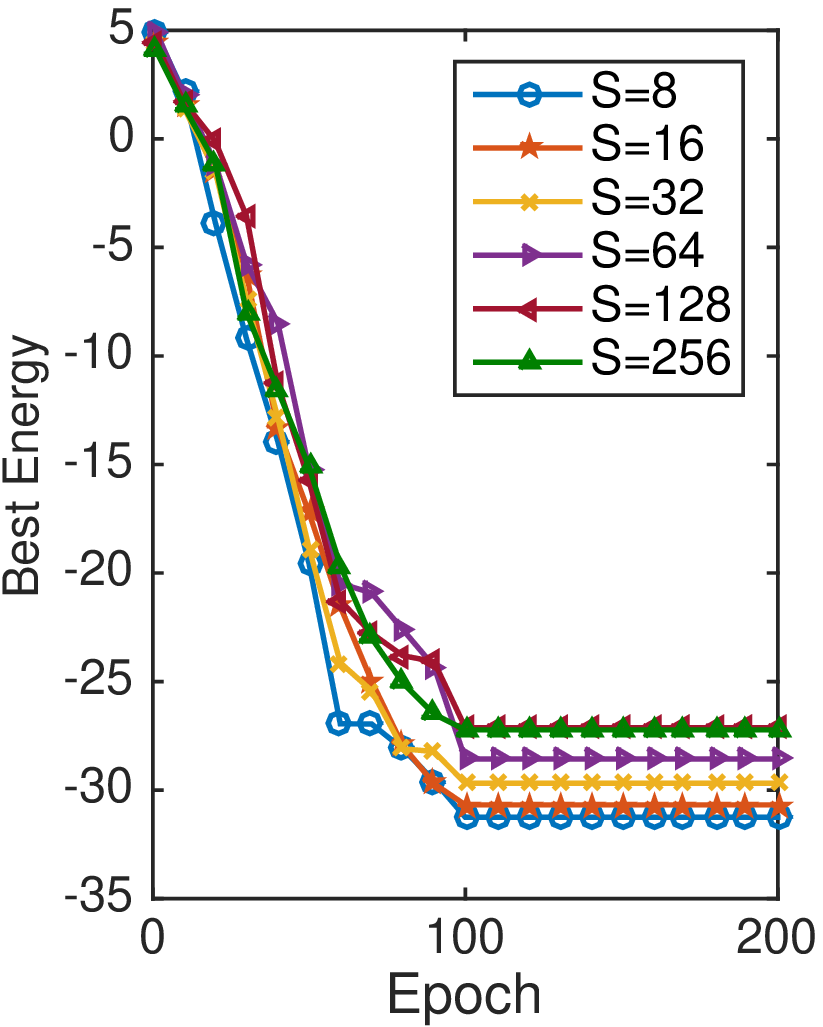}  
\caption{Best energy loss}
\label{fig:radial_sampling_angular}
\end{subfigure}%
\caption{Number of candidate state vectors analysis of \textit{EDropout} for \textbf{ResNet-18} and \textbf{Flowers} dataset.}
\label{fig:pop_EPruning_resenet18_flowers}
\end{minipage}%
~
\begin{minipage}{.49\textwidth}
\centering
\captionsetup{font=footnotesize}
\begin{subfigure}[t]{0.38\textwidth}
\centering
\captionsetup{font=footnotesize}
\includegraphics[width=1\textwidth]{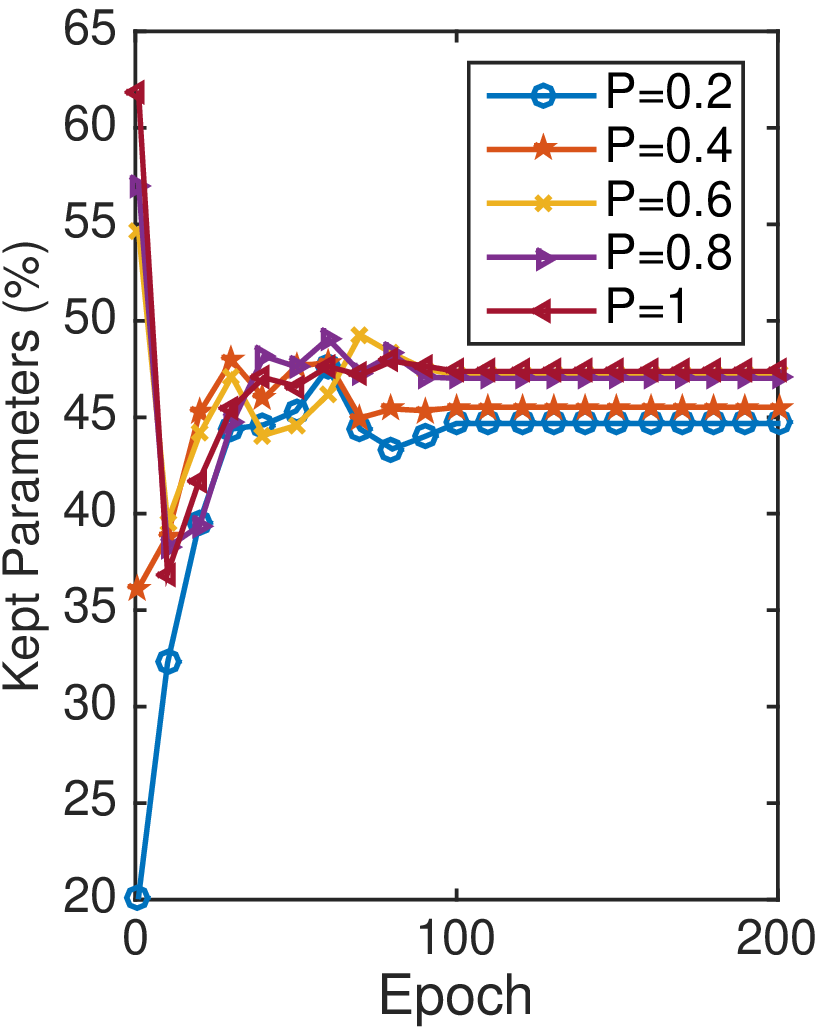}  
\caption{Rate of kept parameters}
\end{subfigure}%
~~~
\begin{subfigure}[t]{0.38\textwidth}
\captionsetup{font=footnotesize}
\centering
\includegraphics[width=1\textwidth]{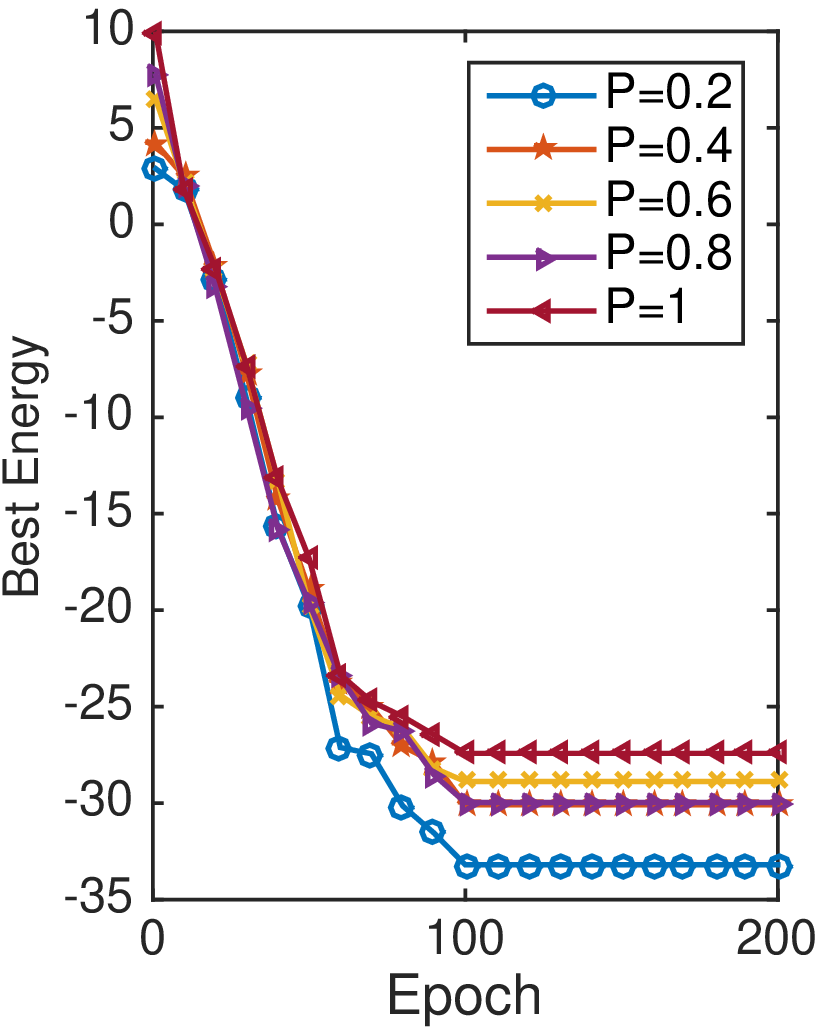}  
\caption{Best energy loss}
\end{subfigure}%
\caption{States initialization analysis of \textit{EDropout} for \textbf{ResNet-18} and \textbf{Flowers} dataset.}
\label{fig:pinit_EPruning_resenet18_flowers}
\end{minipage}
\vspace{-4mm}
\end{figure*}

\section{Experiments}
\label{sec:experiemnts}
We have performed extensive experiments on datasets with different level of difficulty and number of samples on several DNNs with different capacities. First, we analyze the parameters of the \textit{EDropout} method and then discuss the performance results\footnote{The EDropout codes and more details of experiments setup is available at: \textit{https://github.com/sparsifai/epruning}}. 

\subsection{Data}
The following benchmark datasets are used: (i) Fashion (gray images in 10 classes, 54k train, 6k validation, and 10k test)~\cite{xiao2017}, (ii) Kuzushiji (gray images in 10 classes, 54k train, 6k validation, and 10k test)~\cite{clanuwat2018deep}; (iii) CIFAR-10 (color images in 10 classes, 45k train, 5k validation, and 10k test)~\cite{krizhevsky2009learning}, (iv)
CIFAR-100 (color images in 100 classes, 45k train, 5k validation, and 10k test)~\cite{krizhevsky2009learning}, (v) Flowers (102 flower categories; each class has between 40 and 258 images; 10 images from each class for validation and 10 for test)~\cite{nilsback2008automated}, and (vi) ImageNet (color images in 1,000 classes; the training and test datasets are standard from the other implementations\footnote{https://github.com/Eric-mingjie/rethinking-network-pruning})~\cite{deng2009imagenet}. The horizontal flip and Cutout~\cite{devries2017improved} augmentation methods are used for training on CIFAR and Flowers datasets. Input images are resized to $32\times32$ for ResNets, $224\times224$ for AlexNet~\cite{krizhevsky2012imagenet} and SqueezeNet-v1.1~\cite{iandola2016squeezenet}. The input images for all models is of size $224\times224$ for the ImageNet dataset\footnote{https://pytorch.org/tutorials/beginner/transfer\_learning\_tutorial.html}~\cite{deng2009imagenet}.

\subsection{Models}
The experiments are conducted using the following models:
\begin{itemize}
   \item ResNets (18, 34, 50, and 101 layers)\footnote{https://github.com/pytorch/vision/blob/master/torchvision/models/resnet.py}~\cite{he2016deep}.
    \item AlexNet\footnote{https://github.com/pytorch/vision/blob/master/torchvision/models/alexnet.py}~\cite{krizhevsky2012imagenet}, which is computationally very expensive.
    \item SqueezeNet v1.1\footnote{https://github.com/pytorch/vision/blob/master/torchvision/models\\ /squeezenet.py}~\cite{iandola2016squeezenet}, which is a slim design of AlexNet.
    \item Deep Compression\footnote{https://github.com/mightydeveloper/Deep-Compression-PyTorch}~\cite{han2015deep}, which is a pruning and compression method.
    \item $l_{1}$ norm~\cite{li2016pruning} pruning method. It has two versions A and B, where the first has a filter pruning rate of $\approx 10\%$ and skips the sensitive layers 16, 20, 38 and 54 in ResNet-56. The later has a combination three pruning rates for different layers ($60\%$, $30\%$ and $10\%$) and skips layers 16, 18, 20, 34, 38, and 54,~\cite{li2016pruning}. We used version A mainly implemented for ResNet-34 for the ImageNet dataset\footnote{https://github.com/Eric-mingjie/rethinking-network-pruning/tree/master/imagenet/l1-norm-pruning}. 
    \item ThiNet~\cite{luo2017thinet}, which is a filter level pruning method. We use the implementation provided for ResNet-50\footnote{https://github.com/Eric-mingjie/rethinking-network-pruning/tree/master/imagenet/thinet}.
    \item ChannelNet~\cite{gao2018channelnets}, which is a slim network by design. It has three versions and we use ChannelNet-v2\footnote{https://github.com/GenDisc/ChannelNet} which has the highest Top-1 accuracy~\cite{gao2018channelnets} on the ImageNet dataset~\cite{deng2009imagenet}. It also has better performance than MobileNet~\cite{howard2017mobilenets} as reported in~\cite{deng2009imagenet}.
\end{itemize}

\subsection{Training Setup}
 The results are averaged over five independent runs. A grid hyper-parameter search is conducted based on the Top-1 accuracy for all models, including initial learning rates in $\{0.01,0.1,1\}$, Stochastic Gradient Descent (SGD) and Adadelta~\cite{zeiler2012adadelta} optimizer, exponential and step learning rate decays with gamma values in $\{25,50\}$ and batch sizes of 64 and 128. The Adadelta optimizer with Step adaptive learning rate (step: every 50 epochs at gamma rate of 0.1) and weight decay of $10e^{-6}$ is used. The number of epochs is 200 and the batch size is set to 128. Random dropout is not used in the \textit{EDropout} experiments. For the other models, where applicable, the random dropout rate is set to 0.5. The early state convergence in~(\ref{eq:stateconvergence}) is used with a threshold of 100 epochs. The models are implemented in PyTorch and trained using three NVIDIA Titan RTX GPUs.

\subsection{Exploration vs. Exploitation: Dropout Leading to Pruning}
The balance between \textit{exploration} and \textit{exploitation} in searching for the \textit{best state vector} is crucial. The number of candidate state vectors, initialization of the states, mutation factor, and cross-over rate are among the major parameters to control diversity of search.

\begin{table}[!tbp]
\begin{minipage}[t]{\columnwidth}
\centering
\captionsetup{font=footnotesize}
\caption{Performance study of \textbf{ResNet-18} on the \textbf{Flowers} dataset for different early state convergence thresholds $\Delta\mathbf{s}_{T}$.}
\begin{adjustbox}{width=0.75\textwidth}
\begin{tabular}{cccccc}
\hline
$\Delta\mathbf{s}_{T}$ & Loss   & Top-1 & Top-3 & Top-5 & $R$     \\ \hline
50                     & 1.7867 & 55.55 & 77.59 & 83.78 & 48.04 \\ 
100                    & \textbf{1.6550} & \textbf{61.54} & \textbf{79.18} & \textbf{85.55} & 48.19  \\ 
150                    & 1.8088 & 55.55 & 74.86 & 82.12 & 48.78 \\ 
200                    & 1.6853 & 57.31 & 75.73 & 82.89 & \textbf{46.72} \\ \hline
\end{tabular}
\end{adjustbox}
\label{T:earlystop}
\vspace{7mm}
\end{minipage}

\begin{minipage}[t]{\columnwidth}
\centering
\captionsetup{font=footnotesize}
\caption{Performance study of \textbf{ResNet-18} on the \textbf{Flowers} dataset for different number of candidate states $S$.}
\begin{adjustbox}{width=0.75\textwidth}
\begin{tabular}{cccccc}
 \hline
$S$   & Loss   & Top-1   & Top-3   & Top-5   & $R$       \\ \hline
8   & \textbf{1.6550} & \textbf{61.54} & \textbf{79.18} & \textbf{85.55} & 48.19 \\
16  & 1.8029 & 55.93 & 75.74 & 81.92 & 48.60 \\
32  & 1.7523 & 56.45 & 76.43 & 82.91 & 45.96 \\
64  & 1.7248 & 57.20 & 77.69 & 83.97 & 44.72 \\ 
128 & 1.7163 & 57.50 & 76.72 & 82.80 & 42.06 \\
256 & 1.7042 & 58.12 & 77.62 & 83.69 & \textbf{41.69} \\ \hline
\end{tabular}
\end{adjustbox}
\label{T:popsize_resnet18_flowers_EPruning}
\vspace{7mm}
\end{minipage}

\begin{minipage}[t]{\columnwidth}
\centering
\captionsetup{font=footnotesize}
\caption{Performance study of \textbf{ResNet-18} on the \textbf{Flowers} dataset for different states initialization probability $P$.}
\begin{adjustbox}{width=0.75\textwidth}
\begin{tabular}{cccccc}
 \hline
$P$   & Loss   & Top-1   & Top-3   & Top-5   & $R$       \\ \hline
0.2 & 1.6905 & \textbf{60.45} & \textbf{78.51} & \textbf{84.58} & \textbf{44.68} \\ 
0.4 & 1.6776 & 59.39 & 77.60 & 83.80 & 46.25 \\ 
0.6 & 1.7209 & 59.38   & 77.42   & 83.40   & 47.19   \\ 
0.8 & 1.7084 & 58.97   & 77.70   & 83.97   & 47.31   \\ 
1   & \textbf{1.6401} & 59.66 &78.01  & \textbf{84.58}   & 47.38   \\ \hline
\end{tabular}
\end{adjustbox}
\label{T:init_resnet18_flowers_EPruning}
\end{minipage}

\end{table}

\begin{table*}[!ht]
\captionsetup{font=footnotesize}

\caption{Classification performance of ResNets~\cite{he2016deep}, Deep Compression (DC)~\cite{han2015deep}, $l_{1}$ norm~\cite{li2016pruning}, and ThiNet~\cite{takahashi2017aenet} on the test datasets. $R$ is kept trainable parameters and $\#p$ is approximate number of trainable parameters. All the values except loss and $\#p$ are in percentage. (F) refers to full network and (P) refers to pruned network used for inference with EDropout.}
\vspace{6mm}
\begin{subtable}{0.49\linewidth}
\centering
\captionsetup{font=footnotesize}
\caption{\textbf{Kuzushiji}}
\begin{adjustbox}{width=1\textwidth}
\begin{tabular}{lcccccc}
\hline
\multicolumn{1}{c}{Model}  & Loss   & Top-1   & Top-3   & Top-5   & $R$ & $\#p$       \\ \hline \hline
\multicolumn{1}{l}{ResNet-18}                    & \multicolumn{1}{l}{0.0709} 
& \multicolumn{1}{l}{98.23}                    
& \multicolumn{1}{l}{99.53} 
& \multicolumn{1}{l}{99.79} 
& \multicolumn{1}{l}{100}   
& \multicolumn{1}{l}{11.1M}\\ 

\multicolumn{1}{l}{ResNet-18+DC}         
& \multicolumn{1}{l}{0.1617} 
& \multicolumn{1}{l}{95.92}                           
& \multicolumn{1}{l}{98.91} 
& \multicolumn{1}{l}{99.58} 
& \multicolumn{1}{l}{51.49}   
& \multicolumn{1}{l}{5.7M}\\ 

\multicolumn{1}{l}{ResNet-18+EDropout(F)}    
& \multicolumn{1}{l}{0.1112} 
& \multicolumn{1}{l}{97.75}                               & \multicolumn{1}{l}{99.41} 
& \multicolumn{1}{l}{99.78} 
& \multicolumn{1}{l}{100}   
& \multicolumn{1}{l}{11.1M}\\ 

\multicolumn{1}{l}{ResNet-18+EDropout(P)}  
& \multicolumn{1}{l}{0.1107} 
& \multicolumn{1}{l}{97.78}                     
& \multicolumn{1}{l}{99.45} 
& \multicolumn{1}{l}{99.73} 
& \multicolumn{1}{l}{51.49}
&5.7M \\ \hline \hline

\multicolumn{1}{l}{ResNet-34}                    
& \multicolumn{1}{l}{0.0704} 
& \multicolumn{1}{l}{99.52}                     
& \multicolumn{1}{l}{99.90}   
& \multicolumn{1}{l}{99.93} 
& \multicolumn{1}{l}{100} 
& \multicolumn{1}{l}{21.2M}  \\ 

\multicolumn{1}{l}{ResNet-34+$l_{1}$ norm}                    
& \multicolumn{1}{l}{0.1054} 
& \multicolumn{1}{l}{98.50}                     
& \multicolumn{1}{l}{99.02}   
& \multicolumn{1}{l}{99.89} 
& \multicolumn{1}{l}{90.65} 
& \multicolumn{1}{l}{19.2M}  \\

\multicolumn{1}{l}{ResNet-34+DC}       
& \multicolumn{1}{l}{0.2023} 
& \multicolumn{1}{l}{94.42}                           
& \multicolumn{1}{l}{98.66} 
& \multicolumn{1}{l}{99.55}
& \multicolumn{1}{l}{46.11}   
& \multicolumn{1}{l}{9.8M}\\

\multicolumn{1}{l}{ResNet-34+EDropout(F)}    
& \multicolumn{1}{l}{0.1115} 
& \multicolumn{1}{l}{97.78}                               
& \multicolumn{1}{l}{99.42} 
& \multicolumn{1}{l}{99.72} 
& \multicolumn{1}{l}{100}  
& \multicolumn{1}{l}{21.2M}  \\

\multicolumn{1}{l}{ResNet-34+EDropout(P)}  
& \multicolumn{1}{l}{0.1143} 
& \multicolumn{1}{l}{97.71}                           
& \multicolumn{1}{l}{99.44} 
& \multicolumn{1}{l}{99.65} 
& \multicolumn{1}{l}{46.11} 
& \multicolumn{1}{l}{9.8M} \\ \hline \hline

\multicolumn{1}{l}{ResNet-50}                    
& \multicolumn{1}{l}{0.0902} 
& \multicolumn{1}{l}{97.70}                               
&\multicolumn{1}{l}{99.44} 
& \multicolumn{1}{l}{99.79} 
& \multicolumn{1}{l}{100}  
& \multicolumn{1}{l}{23.5M} \\

\multicolumn{1}{l}{ResNet-50+ThiNet}                    
& \multicolumn{1}{l}{0.2278} 
& \multicolumn{1}{l}{95.74}                     
&\multicolumn{1}{l}{97.41} 
& \multicolumn{1}{l}{99.50} 
& \multicolumn{1}{l}{44.04}  
& \multicolumn{1}{l}{10.3M} \\

\multicolumn{1}{l}{ResNet-50+DC}            
& \multicolumn{1}{l}{0.2142} 
& \multicolumn{1}{l}{94.36}                           
& \multicolumn{1}{l}{98.53} 
& \multicolumn{1}{l}{99.36} 
& \multicolumn{1}{l}{45.57}   
& \multicolumn{1}{l}{10.7M} \\

\multicolumn{1}{l}{ResNet-50+EDropout(F)}    
& \multicolumn{1}{l}{0.1250} 
& \multicolumn{1}{l}{97.89}   
&\multicolumn{1}{l}{99.38} 
& \multicolumn{1}{l}{99.75} 
& \multicolumn{1}{l}{100}  
& \multicolumn{1}{l}{23.5M} \\

\multicolumn{1}{l}{ResNet-50+EDropout(P)}  
& \multicolumn{1}{l}{0.1289} 
& \multicolumn{1}{l}{97.63}                               & \multicolumn{1}{l}{99.35} 
& \multicolumn{1}{l}{99.73} 
& \multicolumn{1}{l}{45.57} 
& \multicolumn{1}{l}{10.7M} \\ \hline \hline

\multicolumn{1}{l}{ResNet-101}                   
& \multicolumn{1}{l}{0.0699} 
& \multicolumn{1}{l}{98.26}                               & \multicolumn{1}{l}{99.63} 
& \multicolumn{1}{l}{99.80} 
& \multicolumn{1}{l}{100}  
& \multicolumn{1}{l}{42.5M}   \\

\multicolumn{1}{l}{ResNet-101+DC}             & \multicolumn{1}{l}{0.5648} 
& \multicolumn{1}{l}{93.30}                           
& \multicolumn{1}{l}{98.48} 
& \multicolumn{1}{l}{99.39}
& \multicolumn{1}{l}{46.00}   
& \multicolumn{1}{l}{19.5M}   \\ 

\multicolumn{1}{l}{ResNet-101+EDropout(F)}   
& \multicolumn{1}{l}{0.1140} 
& \multicolumn{1}{l}{98.06}                               & \multicolumn{1}{l}{99.39} 
& \multicolumn{1}{l}{99.69} 
& \multicolumn{1}{l}{100}   
& \multicolumn{1}{l}{42.5M}   \\ 

\multicolumn{1}{l}{ResNet-101+EDropout(P)} 
& \multicolumn{1}{l}{0.1087} 
& \multicolumn{1}{l}{98.05}                               & \multicolumn{1}{l}{99.47}
& \multicolumn{1}{l}{99.69} 
& \multicolumn{1}{l}{46.00}
& \multicolumn{1}{l}{19.5M} \\ \hline

\end{tabular}
\end{adjustbox}
\label{T:results_japan_EPruning}
\end{subtable}
~
\begin{subtable}{0.49\linewidth}
\centering
\captionsetup{font=footnotesize}
\caption{\textbf{Fashion}}
\begin{adjustbox}{width=1\textwidth}
\begin{tabular}{lcccccc}
\hline
\multicolumn{1}{c}{Model}  & Loss   & Top-1   & Top-3   & Top-5   & R & $\#p$      \\ \hline \hline

ResNet-18                    
& 0.2786
& 93.87 
& 99.39 
& 99.68 
& 100   
& \multicolumn{1}{l}{11.1M}\\ 

\multicolumn{1}{l}{ResNet-18+DC}         
& \multicolumn{1}{l}{0.2299} 
& \multicolumn{1}{l}{91.97}                           
& \multicolumn{1}{l}{99.32} 
& \multicolumn{1}{l}{99.86} 
& \multicolumn{1}{l}{50.43}   
& \multicolumn{1}{l}{5.6M} \\

ResNet-18+EDropout(F)    
& 0.4000 
& 93.45
& 99.04 
& 99.47 
& 100  
& \multicolumn{1}{l}{11.1M}\\ 

ResNet-18+EDropout(P) 
& 0.3934
& 93.57 
& 99.06
& 99.47 
& 50.43 
& 5.6M \\ \hline \hline

ResNet-34                   
& 0.3198 
& 93.61 
& 99.12 
& 99.62
& 100  
& \multicolumn{1}{l}{21.2M}  \\ 

\multicolumn{1}{l}{ResNet-34+$l_{1}$ norm}                 
& \multicolumn{1}{l}{0.3742} 
& \multicolumn{1}{l}{91.60}                     
& \multicolumn{1}{l}{97.49}   
& \multicolumn{1}{l}{98.76} 
& \multicolumn{1}{l}{90.65} 
& \multicolumn{1}{l}{19.2M}  \\ 

\multicolumn{1}{l}{ResNet-34+DC}         
& \multicolumn{1}{l}{0.2632} 
& \multicolumn{1}{l}{90.62}                           
& \multicolumn{1}{l}{99.26} 
& \multicolumn{1}{l}{99.87} 
& \multicolumn{1}{l}{46.94}  
& \multicolumn{1}{l}{9.9M}  \\ 

ResNet-34+EDropout(F)    
& 0.4674 
& 92.80 
& 98.78
& 99.35
& 100   
& \multicolumn{1}{l}{21.2M}  \\

ResNet-34+EDropout(P)  
& 0.4582 
& 92.57 
& 98.68
& 99.35 
& 46.94 
& \multicolumn{1}{l}{9.9M}  \\ \hline \hline

ResNet-50                    
&0.3187 
& 93.34 
& 99.15 
& 99.60
& 100  
& \multicolumn{1}{l}{23.5M}  \\ 

\multicolumn{1}{l}{ResNet-50+ThiNet}                   
& \multicolumn{1}{l}{0.3052} 
& \multicolumn{1}{l}{92.19}                               
&\multicolumn{1}{l}{97.93} 
& \multicolumn{1}{l}{98.83} 
& \multicolumn{1}{l}{44.04}  
& \multicolumn{1}{l}{10.3M} \\

\multicolumn{1}{l}{ResNet-50+DC}         
& \multicolumn{1}{l}{0.2956} 
& \multicolumn{1}{l}{89.22}                           
& \multicolumn{1}{l}{98.95} 
& \multicolumn{1}{l}{99.79} 
& \multicolumn{1}{l}{45.14}  
& \multicolumn{1}{l}{10.6M}  \\

ResNet-50+EDropout(F)    
& 0.5451 
& 92.91 
& 98.70 
& 99.34 
& 100  
& \multicolumn{1}{l}{23.5M}  \\

ResNet-50+EDropout(P)  
& 0.5154 
& 92.97 
& 98.79 
& 99.33 
& 45.14
& \multicolumn{1}{l}{10.6M}  \\ 
\hline \hline

ResNet-101                   
&0.3208 
& 93.31 
& 99.10 
& 99.63 
& 100  
& \multicolumn{1}{l}{42.5M}  \\

\multicolumn{1}{l}{ResNet-101+DC}         
& \multicolumn{1}{l}{1.4812} 
& \multicolumn{1}{l}{89.94}                           
& \multicolumn{1}{l}{98.72} 
& \multicolumn{1}{l}{99.46} 
& \multicolumn{1}{l}{45.19}  
& \multicolumn{1}{l}{19.2M}  \\

ResNet-101+EDropout(F)   
& 0.5785 
& 92.58 
& 98.83 
& 99.40 
& 100  
& \multicolumn{1}{l}{42.5M}  \\

ResNet-101+EDropout(P) 
& 0.5717 
& 92.57 
& 98.82 
& 99.35 
& 45.19 
& \multicolumn{1}{l}{19.2M}  \\  \hline

\end{tabular}
\end{adjustbox}
\label{T:results_fashion_EPruning}
\end{subtable}
\vspace{6mm}

\begin{subtable}{0.49\linewidth}
\centering
\captionsetup{font=footnotesize}
\caption{ \textbf{CIFAR-10} }
\begin{adjustbox}{width=1\textwidth}
\begin{tabular}{lcccccc}
\hline
\multicolumn{1}{c}{Model}  & Loss   & Top-1   & Top-3   & Top-5   & $R$ & $\#p$      \\ \hline \hline

\multicolumn{1}{l}{ResNet-18}                    
& \multicolumn{1}{l}{0.3181} 
& \multicolumn{1}{l}{92.81} 
& \multicolumn{1}{l}{98.78} 
& \multicolumn{1}{l}{99.49}
& \multicolumn{1}{l}{100}   
& \multicolumn{1}{l}{11.2M}\\

\multicolumn{1}{l}{ResNet-18+DC}         
& \multicolumn{1}{l}{0.6951} 
& \multicolumn{1}{l}{76.15}                           
& \multicolumn{1}{l}{94.16} 
& \multicolumn{1}{l}{98.59} 
& \multicolumn{1}{l}{49.66} 
& \multicolumn{1}{l}{5.5M}   \\

\multicolumn{1}{l}{ResNet-18+EDropout(F)}        
& \multicolumn{1}{l}{0.4906} 
& \multicolumn{1}{l}{90.96} 
& \multicolumn{1}{l}{98.33} 
& \multicolumn{1}{l}{99.60} 
& \multicolumn{1}{l}{100}  
& \multicolumn{1}{l}{11.2M}\\

\multicolumn{1}{l}{ResNet-18+EDropout(P)}        
& \multicolumn{1}{l}{0.4745} 
& \multicolumn{1}{l}{90.96} 
& \multicolumn{1}{l}{98.40} 
& \multicolumn{1}{l}{99.58} 
& \multicolumn{1}{l}{49.66} 
& \multicolumn{1}{l}{5.5M}\\ \hline\hline

\multicolumn{1}{l}{ResNet-34}                    
& \multicolumn{1}{l}{0.3684} 
& \multicolumn{1}{l}{92.80} 
& \multicolumn{1}{l}{98.85} 
& \multicolumn{1}{l}{99.71} 
& \multicolumn{1}{l}{100}  
& \multicolumn{1}{l}{21.3M}  \\ 

\multicolumn{1}{l}{ResNet-34+$l_{1}$ norm}          
& \multicolumn{1}{l}{0.3888} 
& \multicolumn{1}{l}{90.63}                     
& \multicolumn{1}{l}{97.14}   
& \multicolumn{1}{l}{99.25} 
& \multicolumn{1}{l}{90.65} 
& \multicolumn{1}{l}{19.3M}  \\ 

\multicolumn{1}{l}{ResNet-34+DC}         
& \multicolumn{1}{l}{1.0570} 
& \multicolumn{1}{l}{66.51}                           
& \multicolumn{1}{l}{91.40} 
& \multicolumn{1}{l}{97.68} 
& \multicolumn{1}{l}{38.83}   
& \multicolumn{1}{l}{8.3M}\\

\multicolumn{1}{l}{ResNet-34+EDropou(F)}         
& \multicolumn{1}{l}{0.4576} 
& \multicolumn{1}{l}{88.28} 
& \multicolumn{1}{l}{97.47} 
& \multicolumn{1}{l}{99.31} 
& \multicolumn{1}{l}{100} 
& \multicolumn{1}{l}{21.3M}  \\

\multicolumn{1}{l}{ResNet-34+EDropout(P)}        
& \multicolumn{1}{l}{0.4598} 
& \multicolumn{1}{l}{88.21}   
& \multicolumn{1}{l}{97.48}   
& \multicolumn{1}{l}{99.28}   
& \multicolumn{1}{l}{38.83} 
& \multicolumn{1}{l}{8.3M} \\ \hline\hline

\multicolumn{1}{l}{ResNet-50}                    
& \multicolumn{1}{l}{0.3761} 
& \multicolumn{1}{l}{92.21} 
& \multicolumn{1}{l}{98.70} 
& \multicolumn{1}{l}{99.51} 
& \multicolumn{1}{l}{100}   
& \multicolumn{1}{l}{23.5M} \\

\multicolumn{1}{l}{ResNet-50+ThiNet}                    
& \multicolumn{1}{l}{1.0239} 
& \multicolumn{1}{l}{79.83}                               
&\multicolumn{1}{l}{95.52} 
& \multicolumn{1}{l}{98.38} 
& \multicolumn{1}{l}{43.82}  
& \multicolumn{1}{l}{10.3M} \\

\multicolumn{1}{l}{ResNet-50+DC}         
& \multicolumn{1}{l}{1.0271} 
& \multicolumn{1}{l}{67.53}                           
& \multicolumn{1}{l}{89.92} 
& \multicolumn{1}{l}{96.30} 
& \multicolumn{1}{l}{46.39}   
& \multicolumn{1}{l}{10.9M} \\ 

\multicolumn{1}{l}{ResNet-50+EDropout(F)}       
& \multicolumn{1}{l}{0.6041} 
& \multicolumn{1}{l}{85.22} 
& \multicolumn{1}{l}{96.35} 
& \multicolumn{1}{l}{98.77} 
& \multicolumn{1}{l}{100} 
& \multicolumn{1}{l}{23.5M}   \\

\multicolumn{1}{l}{ResNet-50+EDropout(P)}        
& \multicolumn{1}{l}{0.5953} 
& \multicolumn{1}{l}{85.30}   
& \multicolumn{1}{l}{96.62} 
& \multicolumn{1}{l}{98.76} 
& \multicolumn{1}{l}{46.39} 
& \multicolumn{1}{l}{10.9M} \\ \hline\hline

\multicolumn{1}{l}{ResNet-101}                  
& \multicolumn{1}{l}{0.3680} 
& \multicolumn{1}{l}{92.66} 
& \multicolumn{1}{l}{98.69} 
& \multicolumn{1}{l}{99.65} 
& \multicolumn{1}{l}{100}  
& \multicolumn{1}{l}{42.5M}   \\

\multicolumn{1}{l}{ResNet-101+DC}         
& \multicolumn{1}{l}{1.037} 
& \multicolumn{1}{l}{66.32}                           
& \multicolumn{1}{l}{92.65} 
& \multicolumn{1}{l}{98.11} 
& \multicolumn{1}{l}{45.10}   
& \multicolumn{1}{l}{19.2M} \\ 

\multicolumn{1}{l}{ResNet-101+EDropout(F)}       
& \multicolumn{1}{l}{0.6231} 
& \multicolumn{1}{l}{86.97} 
& \multicolumn{1}{l}{97.42} 
& \multicolumn{1}{l}{99.24} 
& \multicolumn{1}{l}{100}  
& \multicolumn{1}{l}{42.5M}   \\ 

\multicolumn{1}{l}{ResNet-101+EDropout(P)}       
& \multicolumn{1}{l}{0.6339} 
& \multicolumn{1}{l}{86.57} 
& \multicolumn{1}{l}{97.37} 
& \multicolumn{1}{l}{99.20} 
& \multicolumn{1}{l}{45.10} 
& \multicolumn{1}{l}{19.2M} \\ \hline

\end{tabular}
\end{adjustbox}
\label{T:results_cifar10_EPruning}
\end{subtable}
~
\begin{subtable}{0.49\linewidth}
\centering
\captionsetup{font=footnotesize}
\caption{\textbf{CIFAR-100}}
\begin{adjustbox}{width=1\textwidth}
\begin{tabular}{lcccccc}
\hline
\multicolumn{1}{c}{Model}  & Loss   & Top-1   & Top-3   & Top-5   & $R$ & $\#p$      \\ \hline \hline

\multicolumn{1}{l}{ResNet-18}                  
& \multicolumn{1}{l}{1.3830}
& \multicolumn{1}{l}{69.03} 
& \multicolumn{1}{l}{84.44} 
& \multicolumn{1}{l}{88.90} 
& \multicolumn{1}{c}{100}   
& \multicolumn{1}{l}{11.2M}\\

\multicolumn{1}{l}{ResNet-18+DC}         
& \multicolumn{1}{l}{2.3072} 
& \multicolumn{1}{l}{40.01}                           
& \multicolumn{1}{l}{62.20} 
& \multicolumn{1}{l}{72.28} 
& \multicolumn{1}{l}{48.04}  
& \multicolumn{1}{l}{5.4M} \\ 

\multicolumn{1}{l}{ResNet-18+EDropout(F)}        
& \multicolumn{1}{l}{1.9479}
& \multicolumn{1}{l}{67.04} 
& \multicolumn{1}{l}{84.11}   
& \multicolumn{1}{l}{89.43}   
& \multicolumn{1}{c}{100}   
& \multicolumn{1}{l}{11.2M}\\

\multicolumn{1}{l}{ResNet-18+EDropout(P)}    
& \multicolumn{1}{l}{1.9541} 
& \multicolumn{1}{l}{67.06} 
& \multicolumn{1}{l}{84.14} 
& \multicolumn{1}{l}{89.27}   
& \multicolumn{1}{c}{48.04}
& \multicolumn{1}{l}{5.4M}\\  \hline\hline

\multicolumn{1}{l}{ResNet-34}                  
& \multicolumn{1}{l}{1.3931} 
& \multicolumn{1}{l}{69.96} 
& \multicolumn{1}{l}{85.65}  
& \multicolumn{1}{l}{90.10} 
& \multicolumn{1}{l}{100}  
& \multicolumn{1}{l}{21.3M}  \\ 

\multicolumn{1}{l}{ResNet-34+$l_{1}$ norm}                    
& \multicolumn{1}{l}{1.7033} 
& \multicolumn{1}{l}{68.12}                     
& \multicolumn{1}{l}{81.32}   
& \multicolumn{1}{l}{84.73} 
& \multicolumn{1}{l}{90.65} 
& \multicolumn{1}{l}{19.3M}  \\ 

\multicolumn{1}{l}{ResNet-34+DC}         
& \multicolumn{1}{l}{2.1778} 
& \multicolumn{1}{l}{42.09}                           
& \multicolumn{1}{l}{65.01}                           
& \multicolumn{1}{l}{74.31} 
& \multicolumn{1}{l}{49.41} 
& \multicolumn{1}{l}{10.5M}   \\

\multicolumn{1}{l}{ResNet-34+EDropout(F)}     
& \multicolumn{1}{l}{1.9051} 
& \multicolumn{1}{l}{64.50}  
& \multicolumn{1}{l}{81.38}   
& \multicolumn{1}{l}{86.87}   
& \multicolumn{1}{l}{100}   
& \multicolumn{1}{l}{21.3M}  \\

\multicolumn{1}{l}{ResNet-34+EDropout(P)}       
& \multicolumn{1}{l}{1.9219} 
& \multicolumn{1}{l}{64.79}  
& \multicolumn{1}{l}{81.28}  
& \multicolumn{1}{l}{86.74}   
& \multicolumn{1}{c}{49.41}  
& \multicolumn{1}{l}{10.5M}  \\ \hline\hline

\multicolumn{1}{l}{ResNet-50}                    
& \multicolumn{1}{l}{1.3068}
& \multicolumn{1}{l}{71.22}  
& \multicolumn{1}{l}{86.47}  
& \multicolumn{1}{l}{90.74}  
& \multicolumn{1}{c}{100}  
& \multicolumn{1}{l}{23.7M}  \\ 

\multicolumn{1}{l}{ResNet-50+ThiNet}               
& \multicolumn{1}{l}{2.0835} 
& \multicolumn{1}{l}{53.61}                               
&\multicolumn{1}{l}{75.25} 
& \multicolumn{1}{l}{79.73} 
& \multicolumn{1}{l}{44.30}  
& \multicolumn{1}{l}{10.5M} \\

\multicolumn{1}{l}{ResNet-50+DC}         
& \multicolumn{1}{l}{2.3115} 
& \multicolumn{1}{l}{43.87}                           
& \multicolumn{1}{l}{67.02} 
& \multicolumn{1}{l}{76.26} 
& \multicolumn{1}{l}{46.01}  
& \multicolumn{1}{l}{10.9M}  \\

\multicolumn{1}{l}{ResNet-50+EDropout(F)}        
& \multicolumn{1}{l}{1.8750} 
& \multicolumn{1}{l}{61.60}   
& \multicolumn{1}{l}{79.52}   
& \multicolumn{1}{l}{85.45}   
& \multicolumn{1}{c}{100}  
& \multicolumn{1}{l}{23.7M}  \\ 

\multicolumn{1}{l}{ResNet-50+EDropout(P)}    
& \multicolumn{1}{l}{1.8768} 
& \multicolumn{1}{l}{61.91}  
& \multicolumn{1}{l}{79.99}  
& \multicolumn{1}{l}{85.87} 
& \multicolumn{1}{c}{46.01} 
& \multicolumn{1}{l}{10.9M}  \\ \hline\hline 

\multicolumn{1}{l}{ResNet-101}                   
& \multicolumn{1}{l}{1.3574} 
& \multicolumn{1}{l}{71.19}   
& \multicolumn{1}{l}{85.54}   
& \multicolumn{1}{l}{90.00}   
& \multicolumn{1}{c}{100}  
& \multicolumn{1}{c}{42.6M}   \\

\multicolumn{1}{l}{ResNet-101+DC}         
& \multicolumn{1}{l}{2.6003} 
& \multicolumn{1}{l}{37.08}                           
& \multicolumn{1}{l}{58.78} 
& \multicolumn{1}{l}{68.76} 
& \multicolumn{1}{l}{43.76}  
& \multicolumn{1}{c}{18.6M}   \\

\multicolumn{1}{l}{ResNet-101+EDropout(F)}       
& \multicolumn{1}{l}{1.9558} 
& \multicolumn{1}{l}{61.52}   
& \multicolumn{1}{l}{79.71}   
& \multicolumn{1}{l}{85.20}   
& \multicolumn{1}{c}{100} 
& \multicolumn{1}{c}{42.6M}   \\

\multicolumn{1}{l}{ResNet-101+EDropout(P)}       
& \multicolumn{1}{l}{1.9412} 
& \multicolumn{1}{l}{61.92} 
& \multicolumn{1}{l}{79.49} 
& \multicolumn{1}{l}{85.23} 
& \multicolumn{1}{c}{43.76} 
& \multicolumn{1}{c}{18.6M}   \\ \hline

\end{tabular}
\end{adjustbox}
\label{T:results_cifar100_EPruning}
\end{subtable}
\vspace{6mm}

\begin{subtable}{0.49\linewidth}
\centering
\captionsetup{font=footnotesize}
\caption{\textbf{Flowers}}
\begin{adjustbox}{width=1\textwidth}
\begin{tabular}{lcccccc}
\hline
\multicolumn{1}{c}{Model}  & Loss   & Top-1   & Top-3   & Top-5   & $R$ & $\#p$      \\ \hline \hline

\multicolumn{1}{l}{ResNet-18}                    
& \multicolumn{1}{c}{1.8262} 
& \multicolumn{1}{c}{62.60}
& \multicolumn{1}{c}{80.64}
& \multicolumn{1}{c}{86.92} 
& \multicolumn{1}{c}{100}  
& \multicolumn{1}{l}{11.2M}\\

\multicolumn{1}{l}{ResNet-18+DC}         
& \multicolumn{1}{l}{2.4988} 
& \multicolumn{1}{l}{53.92}                           
& \multicolumn{1}{l}{60.68} 
& \multicolumn{1}{l}{76.38} 
& \multicolumn{1}{l}{48.19}   
& \multicolumn{1}{l}{5.4M}\\

\multicolumn{1}{l}{ResNet-18+EDropout(F)}      
& \multicolumn{1}{c}{1.6808} 
& \multicolumn{1}{c}{58.73} 
& \multicolumn{1}{c}{77.19} 
& \multicolumn{1}{c}{82.40} 
& \multicolumn{1}{c}{100}   
& \multicolumn{1}{c}{11.2M}\\

\multicolumn{1}{l}{ResNet-18+EDropout(P)}       
& \multicolumn{1}{c}{1.6550} 
& \multicolumn{1}{c}{61.54}
& \multicolumn{1}{c}{79.18}
& \multicolumn{1}{c}{85.55} 
& \multicolumn{1}{c}{48.19} 
& \multicolumn{1}{l}{5.4M}\\ \hline\hline

\multicolumn{1}{l}{ResNet-34}                
& \multicolumn{1}{c}{1.8993}
& \multicolumn{1}{c}{63.22} 
& \multicolumn{1}{c}{81.08} 
& \multicolumn{1}{c}{87.16}
& \multicolumn{1}{l}{100}   
& \multicolumn{1}{l}{21.3M}  \\ 

\multicolumn{1}{l}{ResNet-34+$l_{1}$ norm}                   
& \multicolumn{1}{l}{1.9935} 
& \multicolumn{1}{l}{60.23}                     
& \multicolumn{1}{l}{83.42}   
& \multicolumn{1}{l}{85.36} 
& \multicolumn{1}{l}{90.65} 
& \multicolumn{1}{l}{19.3M}  \\ 

\multicolumn{1}{l}{ResNet-34+DC}         
& \multicolumn{1}{l}{2.4240} 
& \multicolumn{1}{l}{52.55}                           
& \multicolumn{1}{l}{77.54} 
& \multicolumn{1}{l}{81.46} 
& \multicolumn{1}{l}{42.79}   
& \multicolumn{1}{l}{9.1M}  \\

\multicolumn{1}{l}{ResNet-34+EDropout(F)}       
& \multicolumn{1}{c}{1.6088} 
& \multicolumn{1}{c}{63.96}
& \multicolumn{1}{c}{79.95}
& \multicolumn{1}{c}{85.64}
& \multicolumn{1}{l}{100}  
& \multicolumn{1}{l}{21.3M}  \\

\multicolumn{1}{l}{ResNet-34+EDropout(P)}       
& \multicolumn{1}{c}{1.5960}
& \multicolumn{1}{c}{64.19}
& \multicolumn{1}{c}{80.19}
& \multicolumn{1}{c}{85.58} 
& \multicolumn{1}{l}{42.79} 
& \multicolumn{1}{l}{9.1M}  \\ \hline \hline

\multicolumn{1}{l}{ResNet-50}      
& \multicolumn{1}{c}{2.4766}
& \multicolumn{1}{c}{63.75} 
& \multicolumn{1}{c}{80.24} 
& \multicolumn{1}{c}{87.21} 
& \multicolumn{1}{l}{100}  
& \multicolumn{1}{l}{23.7M}   \\

\multicolumn{1}{l}{ResNet-50+ThiNet}                    
& \multicolumn{1}{l}{2.3883} 
& \multicolumn{1}{l}{41.76}                               
&\multicolumn{1}{l}{66.29} 
& \multicolumn{1}{l}{72.41} 
& \multicolumn{1}{l}{44.30}  
& \multicolumn{1}{l}{10.5M} \\

\multicolumn{1}{l}{ResNet-50+DC}         
& \multicolumn{1}{l}{3.0556} 
& \multicolumn{1}{l}{27.28}                           
& \multicolumn{1}{l}{47.43} 
& \multicolumn{1}{l}{57.63} 
& \multicolumn{1}{l}{44.68}   
& \multicolumn{1}{l}{10.6M}   \\

\multicolumn{1}{l}{ResNet-50+EDropout(F)}   
&\multicolumn{1}{c}{1.8492}      
&\multicolumn{1}{c}{54.60}       
&\multicolumn{1}{c}{76.46}       
&\multicolumn{1}{c}{82.84}       
&\multicolumn{1}{l}{100}   
& \multicolumn{1}{l}{23.7M}   \\

\multicolumn{1}{l}{ResNet-50+EDropout(P)}       
&\multicolumn{1}{c}{1.8293}      
&\multicolumn{1}{c}{56.02}      
&\multicolumn{1}{c}{76.42}      
&\multicolumn{1}{c}{83.10}     
&\multicolumn{1}{l}{44.68}        
& \multicolumn{1}{l}{10.6M}   \\ \hline\hline

\multicolumn{1}{l}{ResNet-101}                  
& \multicolumn{1}{c}{2.6183} 
& \multicolumn{1}{c}{62.70} 
& \multicolumn{1}{c}{82.04}   
& \multicolumn{1}{l}{86.26}   
& \multicolumn{1}{l}{100}   
& \multicolumn{1}{l}{42.7M}   \\ 

\multicolumn{1}{l}{ResNet-101+DC}         
& \multicolumn{1}{l}{3.0623} 
& \multicolumn{1}{l}{28.36}                           
& \multicolumn{1}{l}{46.78} 
& \multicolumn{1}{l}{56.32} 
& \multicolumn{1}{l}{44.85}  
& \multicolumn{1}{l}{19.2M} \\ 

\multicolumn{1}{l}{ResNet-101+EDropout(F)}      
& \multicolumn{1}{c}{1.8241} 
& \multicolumn{1}{c}{59.44}
& \multicolumn{1}{c}{79.02}
& \multicolumn{1}{c}{85.67}  
& \multicolumn{1}{l}{100}  
& \multicolumn{1}{l}{42.7M}   \\ 
\multicolumn{1}{l}{ResNet-101+EDropout(P)}     
& \multicolumn{1}{c}{1.8575} 
& \multicolumn{1}{c}{58.17} 
& \multicolumn{1}{c}{78.32} 
& \multicolumn{1}{c}{85.10}  
& \multicolumn{1}{l}{44.85}   
& \multicolumn{1}{l}{19.2M}   \\ \hline 

\end{tabular}
\end{adjustbox}
\label{T:results_flowers_EPruning}
\end{subtable} 
~
\begin{subtable}{0.49\linewidth}
\centering
\captionsetup{font=footnotesize}
\caption{\textbf{ImageNet}}
\begin{adjustbox}{width=1\textwidth}
\begin{tabular}{lcccccc}
\hline
\multicolumn{1}{c}{Model}  & Loss   & Top-1   & Top-3   & Top-5   & $R$ & $\#p$      \\ \hline \hline

\multicolumn{1}{l}{ResNet-18}                    
& \multicolumn{1}{c}{1.9033} 
& \multicolumn{1}{c}{68.48}
& \multicolumn{1}{c}{81.84}
& \multicolumn{1}{c}{89.01} 
& \multicolumn{1}{c}{100}  
& \multicolumn{1}{l}{11.5M}\\

\multicolumn{1}{l}{ResNet-18+DC}         
& \multicolumn{1}{l}{2.4053} 
& \multicolumn{1}{l}{55.03}                         
& \multicolumn{1}{l}{70.23} 
& \multicolumn{1}{l}{77.14} 
& \multicolumn{1}{l}{46.27}   
& \multicolumn{1}{l}{5.3M}\\

\multicolumn{1}{l}{ResNet-18+EDropout(F)}      
& \multicolumn{1}{c}{1.8503}
& \multicolumn{1}{c}{64.95} 
& \multicolumn{1}{c}{82.23} 
& \multicolumn{1}{c}{87.16}
& \multicolumn{1}{c}{100}   
& \multicolumn{1}{c}{11.5M}\\

\multicolumn{1}{l}{ResNet-18+EDropout(P)}       
& \multicolumn{1}{c}{1.8550} 
& \multicolumn{1}{c}{65.43}
& \multicolumn{1}{c}{82.53}
& \multicolumn{1}{c}{87.26} 
& \multicolumn{1}{c}{46.27} 
& \multicolumn{1}{l}{5.3M}\\ \hline\hline

\multicolumn{1}{l}{ResNet-34}                
& \multicolumn{1}{c}{1.4921}
& \multicolumn{1}{c}{73.42} 
& \multicolumn{1}{c}{85.93} 
& \multicolumn{1}{c}{91.26}
& \multicolumn{1}{l}{100}   
& \multicolumn{1}{l}{21.6M}  \\ 

\multicolumn{1}{l}{ResNet-34+$l_{1}$ norm}         
& \multicolumn{1}{l}{1.5294} 
& \multicolumn{1}{l}{72.42}                     
& \multicolumn{1}{l}{87.29}   
& \multicolumn{1}{l}{90.83} 
& \multicolumn{1}{l}{89.35} 
& \multicolumn{1}{l}{19.3M}  \\ 

\multicolumn{1}{l}{ResNet-34+DC}         
& \multicolumn{1}{l}{1.9606} 
& \multicolumn{1}{l}{63.71}                         
& \multicolumn{1}{l}{81.15} 
& \multicolumn{1}{l}{88.03} 
& \multicolumn{1}{l}{44.06}   
& \multicolumn{1}{l}{9.5M}  \\

\multicolumn{1}{l}{ResNet-34+EDropout(F)}       
& \multicolumn{1}{c}{1.4148} 
& \multicolumn{1}{c}{71.39}
& \multicolumn{1}{c}{86.93}
& \multicolumn{1}{c}{90.27}
& \multicolumn{1}{l}{100}  
& \multicolumn{1}{l}{21.6M}  \\

\multicolumn{1}{l}{ResNet-34+EDropout(P)}       
& \multicolumn{1}{c}{1.3996}
& \multicolumn{1}{c}{71.82}
& \multicolumn{1}{c}{87.20}
& \multicolumn{1}{c}{90.96} 
& \multicolumn{1}{l}{44.06} 
& \multicolumn{1}{l}{9.5M}  \\ \hline \hline

\multicolumn{1}{l}{ResNet-50}      
& \multicolumn{1}{c}{1.1959}
& \multicolumn{1}{c}{75.27} 
& \multicolumn{1}{c}{87.48} 
& \multicolumn{1}{c}{91.87} 
& \multicolumn{1}{l}{100}  
& \multicolumn{1}{l}{24M}   \\

\multicolumn{1}{l}{ResNet-50+ThiNet}                   
& \multicolumn{1}{l}{1.2303} 
& \multicolumn{1}{l}{72.11}                         
&\multicolumn{1}{l}{86.30} 
& \multicolumn{1}{l}{90.37} 
& \multicolumn{1}{l}{51.66}  
& \multicolumn{1}{l}{12.4M} \\

\multicolumn{1}{l}{ResNet-50+DC}         
& \multicolumn{1}{l}{2.0390} 
& \multicolumn{1}{l}{67.02}                         
& \multicolumn{1}{l}{83.53} 
& \multicolumn{1}{l}{86.67} 
& \multicolumn{1}{l}{42.74}   
& \multicolumn{1}{l}{10.2M}   \\

\multicolumn{1}{l}{ResNet-50+EDropout(F)}   
&\multicolumn{1}{c}{1.1905}      
&\multicolumn{1}{c}{73.06}       
&\multicolumn{1}{c}{87.09}       
&\multicolumn{1}{c}{90.19}       
&\multicolumn{1}{l}{100}   
& \multicolumn{1}{l}{24M}   \\

\multicolumn{1}{l}{ResNet-50+EDropout(P)}       
&\multicolumn{1}{c}{1.0502}      
&\multicolumn{1}{c}{73.72}      
&\multicolumn{1}{c}{88.21}      
&\multicolumn{1}{c}{91.21}     
&\multicolumn{1}{l}{42.74}        
& \multicolumn{1}{l}{10.2M}   \\ \hline\hline

\multicolumn{1}{l}{ResNet-101}                  
& \multicolumn{1}{c}{1.5821} 
& \multicolumn{1}{c}{75.94} 
& \multicolumn{1}{c}{86.34}   
& \multicolumn{1}{l}{92.41}   
& \multicolumn{1}{l}{100}   
& \multicolumn{1}{l}{43M}   \\ 

\multicolumn{1}{l}{ResNet-101+DC}         
& \multicolumn{1}{l}{2.1014} 
& \multicolumn{1}{l}{64.62}                     
& \multicolumn{1}{l}{79.28} 
& \multicolumn{1}{l}{85.24} 
& \multicolumn{1}{l}{42.94}  
& \multicolumn{1}{l}{18.46M} \\ 

\multicolumn{1}{l}{ResNet-101+EDropout(F)}      
& \multicolumn{1}{c}{1.6391} 
& \multicolumn{1}{c}{72.52} 
& \multicolumn{1}{c}{85.33}   
& \multicolumn{1}{l}{91.21} 
& \multicolumn{1}{l}{100}   
& \multicolumn{1}{l}{43M}   \\ 

\multicolumn{1}{l}{ResNet-101+EDropout(P)}     
& \multicolumn{1}{c}{1.5519} 
& \multicolumn{1}{c}{73.94} 
& \multicolumn{1}{c}{87.34}   
& \multicolumn{1}{l}{92.08} 
& \multicolumn{1}{l}{42.94}  
& \multicolumn{1}{l}{18.46M} \\ \hline

\end{tabular}
\end{adjustbox}
\label{T:results_imagenet_EPruning}
\end{subtable} 
\label{T:results_EPruning}
\end{table*}

\begin{table*}[!ht]
\captionsetup{font=footnotesize}

\caption{Classification performance of ChannelNet~\cite{gao2018channelnets}, AlexNet~\cite{krizhevsky2012imagenet}, and SqueezeNet~\cite{iandola2016squeezenet} on the test datasets. $R$ is kept trainable parameters and $\#p$ is approximate number of trainable parameters. All the values except loss and $\#p$ are in percentage. (F) refers to full network used for inference and (P) refers to pruned network using \textit{EDropout}. ChannelNet and SqueezeNet are compact models, which is a different approach from pruning.}

\begin{subtable}{0.49\linewidth}
\centering
\captionsetup{font=footnotesize}
\caption{\textbf{Kuzushiji}}
\begin{adjustbox}{width=1\textwidth}
\begin{tabular}{lcccccc}
\hline
\multicolumn{1}{c}{Model}  & Loss   & Top-1   & Top-3   & Top-5   & $R$ & $\#p$       \\ \hline \hline

\multicolumn{1}{l}{ChannelNet-v2}       
& \multicolumn{1}{l}{0.0788} 
& \multicolumn{1}{l}{98.15} 
& \multicolumn{1}{l}{99.56} 
& \multicolumn{1}{l}{99.89} 
& \multicolumn{1}{l}{100} 
& \multicolumn{1}{l}{1.8M} \\ 

\multicolumn{1}{l}{AlexNet}       
& \multicolumn{1}{l}{0.1162} 
& \multicolumn{1}{l}{97.77} 
& \multicolumn{1}{l}{99.43} 
& \multicolumn{1}{l}{99.85} 
& \multicolumn{1}{l}{100} 
& \multicolumn{1}{l}{57M} \\

\multicolumn{1}{l}{AlexNet+EDropout(F)}       
& \multicolumn{1}{l}{0.1976} 
& \multicolumn{1}{l}{96.53} 
& \multicolumn{1}{l}{99.25} 
& \multicolumn{1}{l}{99.72} 
& \multicolumn{1}{l}{100} 
& \multicolumn{1}{l}{57M} \\ 

\multicolumn{1}{l}{AlexNet+EDropout(P)}       
& \multicolumn{1}{l}{0.2089} 
& \multicolumn{1}{l}{96.57} 
& \multicolumn{1}{l}{99.24} 
& \multicolumn{1}{l}{99.68} 
& \multicolumn{1}{l}{78.57}
& \multicolumn{1}{l}{44.8M} \\ 

\multicolumn{1}{l}{SqueezeNet}       
& \multicolumn{1}{l}{0.2114} 
& \multicolumn{1}{l}{97.19} 
& \multicolumn{1}{l}{99.27} 
& \multicolumn{1}{l}{99.72} 
& \multicolumn{1}{l}{100}
& \multicolumn{1}{l}{0.72M} \\

\multicolumn{1}{l}{SqueezeNet+EDropout(F)}       
& \multicolumn{1}{l}{0.2414} 
& \multicolumn{1}{l}{96.45} 
& \multicolumn{1}{l}{99.05} 
& \multicolumn{1}{l}{99.57} 
& \multicolumn{1}{l}{100} 
& \multicolumn{1}{l}{0.72M} \\

\multicolumn{1}{l}{SqueezeNet+EDropout(P)}       
& \multicolumn{1}{l}{0.2411} 
& \multicolumn{1}{l}{96.35} 
& \multicolumn{1}{l}{98.91} 
& \multicolumn{1}{l}{99.54} 
& \multicolumn{1}{l}{49.86} 
& \multicolumn{1}{l}{0.36M} \\ \hline

\end{tabular}
\end{adjustbox}
\label{T:results_japan_EPruning_alexnet}
\end{subtable}
~
\begin{subtable}{0.49\linewidth}
\centering
\captionsetup{font=footnotesize}
\caption{\textbf{Fashion}}
\begin{adjustbox}{width=1\textwidth}
\begin{tabular}{lcccccc}
\hline
\multicolumn{1}{c}{Model}  & Loss   & Top-1   & Top-3   & Top-5   & R & $\#p$      \\ \hline \hline

\multicolumn{1}{l}{ChannelNet-v2}       
& \multicolumn{1}{l}{0.2442} 
& \multicolumn{1}{l}{93.41} 
& \multicolumn{1}{l}{98.03} 
& \multicolumn{1}{l}{99.84} 
& \multicolumn{1}{l}{100} 
& \multicolumn{1}{l}{1.7M} \\ 

\multicolumn{1}{l}{AlexNet}       
& \multicolumn{1}{l}{0.4441} 
& \multicolumn{1}{l}{92.87} 
& \multicolumn{1}{l}{99.27} 
& \multicolumn{1}{l}{99.70} 
& \multicolumn{1}{l}{100} 
& \multicolumn{1}{l}{57M}  \\

\multicolumn{1}{l}{AlexNet+EDropout(F)}       
& \multicolumn{1}{l}{0.3726} 
& \multicolumn{1}{l}{91.21} 
& \multicolumn{1}{l}{99.25} 
& \multicolumn{1}{l}{99.87} 
& \multicolumn{1}{l}{100} 
& \multicolumn{1}{l}{57M}  \\

\multicolumn{1}{l}{AlexNet+EDropout(P)}       
& \multicolumn{1}{l}{0.3862} 
& \multicolumn{1}{l}{91.19} 
& \multicolumn{1}{l}{99.21} 
& \multicolumn{1}{l}{99.86} 
& \multicolumn{1}{l}{77.58}
& \multicolumn{1}{l}{44.2M}  \\

\multicolumn{1}{l}{SqueezeNet}       
& \multicolumn{1}{l}{0.3655} 
& \multicolumn{1}{l}{92.64} 
& \multicolumn{1}{l}{99.48} 
& \multicolumn{1}{l}{99.90} 
& \multicolumn{1}{l}{100}
& \multicolumn{1}{l}{0.72M}  \\

\multicolumn{1}{l}{SqueezeNet+EDropout(F)}       
& \multicolumn{1}{l}{0.2524} 
& \multicolumn{1}{l}{92.25} 
& \multicolumn{1}{l}{99.35} 
& \multicolumn{1}{l}{99.89} 
& \multicolumn{1}{l}{100} 
& \multicolumn{1}{l}{0.72M}  \\

\multicolumn{1}{l}{SqueezeNet+EDropout(P)}       
& \multicolumn{1}{l}{0.2478} 
& \multicolumn{1}{l}{92.34} 
& \multicolumn{1}{l}{99.42} 
& \multicolumn{1}{l}{99.88} 
& \multicolumn{1}{l}{52.83} 
& \multicolumn{1}{l}{0.38M}  \\ 
\hline
\end{tabular}
\end{adjustbox}
\label{T:results_fashion_EPruning_alexnet}
\end{subtable}
\vspace{3mm}

\begin{subtable}{0.49\linewidth}
\centering
\captionsetup{font=footnotesize}
\caption{ \textbf{CIFAR-10} }
\begin{adjustbox}{width=1\textwidth}
\begin{tabular}{lcccccc}
\hline
\multicolumn{1}{c}{Model}  & Loss   & Top-1   & Top-3   & Top-5   & $R$ & $\#p$      \\ \hline \hline

\multicolumn{1}{l}{ChannelNet-v2}       
& \multicolumn{1}{l}{0.3751} 
& \multicolumn{1}{l}{92.5} 
& \multicolumn{1}{l}{98.48} 
& \multicolumn{1}{l}{99.14} 
& \multicolumn{1}{l}{100} 
& \multicolumn{1}{l}{1.7M} \\ 

\multicolumn{1}{l}{AlexNet}       
& \multicolumn{1}{l}{0.9727} 
& \multicolumn{1}{l}{84.32} 
& \multicolumn{1}{l}{96.58} 
& \multicolumn{1}{l}{99.08} 
& \multicolumn{1}{l}{100} 
& \multicolumn{1}{l}{57.4M} \\

\multicolumn{1}{l}{AlexNet+EDropout(F)}       
& \multicolumn{1}{l}{0.7632} 
& \multicolumn{1}{l}{75.05} 
& \multicolumn{1}{l}{93.74} 
& \multicolumn{1}{l}{98.18} 
& \multicolumn{1}{l}{100} 
& \multicolumn{1}{l}{57.4M} \\

\multicolumn{1}{l}{AlexNet+EDropout(P)}       
& \multicolumn{1}{l}{0.7897} 
& \multicolumn{1}{l}{74.66} 
& \multicolumn{1}{l}{93.63} 
& \multicolumn{1}{l}{97.96} 
& \multicolumn{1}{l}{77.36} 
& \multicolumn{1}{l}{44.4M} \\

\multicolumn{1}{l}{SqueezeNet}       
& \multicolumn{1}{l}{0.5585} 
& \multicolumn{1}{l}{81.49} 
& \multicolumn{1}{l}{96.31} 
& \multicolumn{1}{l}{99.01} 
& \multicolumn{1}{l}{100} 
& \multicolumn{1}{l}{0.73M} \\

\multicolumn{1}{l}{SqueezeNet+EDropout(F)}       
& \multicolumn{1}{l}{0.6686} 
& \multicolumn{1}{l}{76.76} 
& \multicolumn{1}{l}{94.55} 
& \multicolumn{1}{l}{98.62} 
& \multicolumn{1}{l}{100} 
& \multicolumn{1}{l}{0.73M} \\

\multicolumn{1}{l}{SqueezeNet+EDropout(P)}       
& \multicolumn{1}{l}{0.6725} 
& \multicolumn{1}{l}{76.85} 
& \multicolumn{1}{l}{95.00} 
& \multicolumn{1}{l}{98.56} 
& \multicolumn{1}{l}{52.35}
& \multicolumn{1}{l}{0.38M} \\  \hline
\end{tabular}
\end{adjustbox}
\label{T:results_cifar10_EPruning_alexnet}
\end{subtable}
~
\begin{subtable}{0.49\linewidth}
\centering
\captionsetup{font=footnotesize}
\caption{\textbf{CIFAR-100}}
\begin{adjustbox}{width=1\textwidth}
\begin{tabular}{lcccccc}
\hline
\multicolumn{1}{c}{Model}  & Loss   & Top-1   & Top-3   & Top-5   & $R$ & $\#p$      \\ \hline \hline

\multicolumn{1}{l}{ChannelNet-v2}       
& \multicolumn{1}{l}{1.8124} 
& \multicolumn{1}{l}{69.7} 
& \multicolumn{1}{l}{83.82} 
& \multicolumn{1}{l}{85.25} 
& \multicolumn{1}{l}{100} 
& \multicolumn{1}{l}{1.8M} \\ 

\multicolumn{1}{l}{AlexNet}       
& \multicolumn{1}{l}{2.8113} 
& \multicolumn{1}{l}{60.12} 
& \multicolumn{1}{l}{79.18} 
& \multicolumn{1}{l}{83.31} 
& \multicolumn{1}{l}{100} 
& \multicolumn{1}{l}{57.4M} \\

\multicolumn{1}{l}{AlexNet+EDropout(F)}       
& \multicolumn{1}{l}{2.4731} 
& \multicolumn{1}{l}{56.62} 
& \multicolumn{1}{l}{78.72} 
& \multicolumn{1}{l}{81.92} 
& \multicolumn{1}{l}{100} 
& \multicolumn{1}{l}{57.4M} \\

\multicolumn{1}{l}{AlexNet+EDropout(P)}       
& \multicolumn{1}{l}{2.4819} 
& \multicolumn{1}{l}{56.59} 
& \multicolumn{1}{l}{78.52} 
& \multicolumn{1}{l}{81.62} 
& \multicolumn{1}{l}{71.84} 
& \multicolumn{1}{l}{41.2M} \\

\multicolumn{1}{l}{SqueezeNet}       
& \multicolumn{1}{l}{1.4150} 
& \multicolumn{1}{l}{67.85} 
& \multicolumn{1}{l}{85.81} 
& \multicolumn{1}{l}{89.69} 
& \multicolumn{1}{l}{100}
& \multicolumn{1}{l}{0.77M} \\ 

\multicolumn{1}{l}{SqueezeNet+EDropout(F)}       
& \multicolumn{1}{l}{1.5265} 
& \multicolumn{1}{l}{64.23} 
& \multicolumn{1}{l}{82.71} 
& \multicolumn{1}{l}{88.63} 
& \multicolumn{1}{l}{100}
& \multicolumn{1}{l}{0.77M} \\ 

\multicolumn{1}{l}{SqueezeNet+EDropout(P)}       
& \multicolumn{1}{l}{1.5341} 
& \multicolumn{1}{l}{64.02} 
& \multicolumn{1}{l}{81.63} 
& \multicolumn{1}{l}{88.51} 
& \multicolumn{1}{l}{56.40} 
& \multicolumn{1}{l}{0.43M} \\ \hline

\end{tabular}
\end{adjustbox}
\label{T:results_cifar100_EPruning_alexnet}
\end{subtable}
\vspace{5mm}

\begin{subtable}{0.49\linewidth}
\centering
\captionsetup{font=footnotesize}
\caption{\textbf{Flowers}}
\begin{adjustbox}{width=1\textwidth}
\begin{tabular}{lcccccc}
\hline
\multicolumn{1}{c}{Model}  & Loss   & Top-1   & Top-3   & Top-5   & $R$ & $\#p$      \\ \hline \hline

\multicolumn{1}{l}{ChannelNet-v2}       
& \multicolumn{1}{l}{2.7517} 
& \multicolumn{1}{l}{63.78} 
& \multicolumn{1}{l}{84.62} 
& \multicolumn{1}{l}{87.13} 
& \multicolumn{1}{l}{100} 
& \multicolumn{1}{l}{1.8M} \\ 

\multicolumn{1}{l}{AlexNet}       
& \multicolumn{1}{l}{2.6872} 
& \multicolumn{1}{l}{56.11} 
& \multicolumn{1}{l}{74.85} 
& \multicolumn{1}{l}{81.92} 
& \multicolumn{1}{l}{100} 
& \multicolumn{1}{l}{57.4M}\\ 
\multicolumn{1}{l}{AlexNet+EDropout(F)}       
& \multicolumn{1}{l}{2.5272} 
& \multicolumn{1}{l}{51.54} 
& \multicolumn{1}{l}{70.78} 
& \multicolumn{1}{l}{80.92} 
& \multicolumn{1}{l}{100}
& \multicolumn{1}{l}{57.4M}\\ 
\multicolumn{1}{l}{AlexNet+EDropout(P)}       
& \multicolumn{1}{l}{2.5159} 
& \multicolumn{1}{l}{51.79} 
& \multicolumn{1}{l}{71.12} 
& \multicolumn{1}{l}{80.67} 
& \multicolumn{1}{l}{81.12} 
& \multicolumn{1}{l}{46.5M}\\ 
\multicolumn{1}{l}{SqueezeNet}       
& \multicolumn{1}{l}{2.2842} 
& \multicolumn{1}{l}{45.11} 
& \multicolumn{1}{l}{63.66} 
& \multicolumn{1}{l}{72.51} 
& \multicolumn{1}{l}{100}
& \multicolumn{1}{l}{0.77M}\\ 
\multicolumn{1}{l}{SqueezeNet+EDropout(F)}       
& \multicolumn{1}{l}{2.2217} 
& \multicolumn{1}{l}{42.76} 
& \multicolumn{1}{l}{62.90} 
& \multicolumn{1}{l}{72.02} 
& \multicolumn{1}{l}{100} 
& \multicolumn{1}{l}{0.77M}\\ 
\multicolumn{1}{l}{SqueezeNet+EDropout(P)}       
& \multicolumn{1}{l}{2.2128} 
& \multicolumn{1}{l}{42.89} 
& \multicolumn{1}{l}{62.80} 
& \multicolumn{1}{l}{72.90} 
& \multicolumn{1}{l}{74.48} 
& \multicolumn{1}{l}{0.57M}\\ \hline
\end{tabular}
\end{adjustbox}
\label{T:results_flowers_EPruning_alexnet}
\end{subtable} 
~
\begin{subtable}{0.49\linewidth}
\centering
\captionsetup{font=footnotesize}
\caption{\textbf{ImageNet}}
\begin{adjustbox}{width=1\textwidth}
\begin{tabular}{lcccccc}
\hline
\multicolumn{1}{c}{Model}  & Loss   & Top-1   & Top-3   & Top-5   & $R$ & $\#p$      \\ \hline \hline

\multicolumn{1}{l}{ChannelNet-v2}       
& \multicolumn{1}{l}{1.4204} 
& \multicolumn{1}{l}{69.5} 
& \multicolumn{1}{l}{87.73} 
& \multicolumn{1}{l}{89.26} 
& \multicolumn{1}{l}{100} 
& \multicolumn{1}{l}{2.7M} \\ 

\multicolumn{1}{l}{AlexNet}       
& \multicolumn{1}{l}{2.6872} 
& \multicolumn{1}{l}{58.26} 
& \multicolumn{1}{l}{79.48} 
& \multicolumn{1}{l}{81.16} 
& \multicolumn{1}{l}{100} 
& \multicolumn{1}{l}{60M}\\ 

\multicolumn{1}{l}{AlexNet+EDropout(F)}       
& \multicolumn{1}{l}{2.1502} 
& \multicolumn{1}{l}{55.39} 
& \multicolumn{1}{l}{77.52} 
& \multicolumn{1}{l}{80.84} 
& \multicolumn{1}{l}{100}
& \multicolumn{1}{l}{60M}\\ 

\multicolumn{1}{l}{AlexNet+EDropout(P)}       
& \multicolumn{1}{l}{2.0582} 
& \multicolumn{1}{l}{56.03} 
& \multicolumn{1}{l}{77.92} 
& \multicolumn{1}{l}{81.15} 
& \multicolumn{1}{l}{82.04} 
& \multicolumn{1}{l}{49.2M}\\ 

\multicolumn{1}{l}{SqueezeNet}       
& \multicolumn{1}{l}{1.5042} 
& \multicolumn{1}{l}{57.31} 
& \multicolumn{1}{l}{78.93} 
& \multicolumn{1}{l}{81.11} 
& \multicolumn{1}{l}{100}
& \multicolumn{1}{l}{1.3M}\\ 

\multicolumn{1}{l}{SqueezeNet+EDropout(F)}       
& \multicolumn{1}{l}{1.6102} 
& \multicolumn{1}{l}{56.04} 
& \multicolumn{1}{l}{76.27} 
& \multicolumn{1}{l}{79.43} 
& \multicolumn{1}{l}{100} 
& \multicolumn{1}{l}{1.3M}\\ 

\multicolumn{1}{l}{SqueezeNet+EDropout(P)}       
& \multicolumn{1}{l}{1.5961} 
& \multicolumn{1}{l}{56.61} 
& \multicolumn{1}{l}{77.01} 
& \multicolumn{1}{l}{79.84} 
& \multicolumn{1}{l}{60.32} 
& \multicolumn{1}{l}{0.78M}\\ \hline
\end{tabular}
\end{adjustbox}
\label{T:results_imagenet_EPruning_alexnet}
\end{subtable} 
\label{T:results_EPruning_alexnet}
\end{table*}

\subsubsection{Early State Convergence}
\label{sec:exp_earlystateconvergence}
Balancing the exploration of optimizer for a feasible state vector while giving enough time for fine-tuning that state is crucial. However, \textit{EDropout} combines the state selection and training in the first phase and ultimately can naturally converge to a \textit{best state vector} based on the energy of the states or manually be controlled by $\Delta\mathbf{s}_{T}$. Figure~\ref{fig:threshold_earlystate_resenet18_flowers}(a) shows convergence of the kept number
of parameters for $\Delta\mathbf{s}_{T}\in\{50, 100, 150, 200\}$. For $\Delta\mathbf{s}_{T}=150$ and $\Delta\mathbf{s}_{T}=200$ the model has converged
approximately at epochs 125 and 160, respectively. Figure~\ref{fig:threshold_earlystate_resenet18_flowers}(c) shows the value of $\Delta\mathbf{s}$ defined in~(\ref{eq:stateconvergence}), whereas $|\Delta\mathbf{s}|\rightarrow 0$ suggests
the algorithm is moving toward \textit{exploitation} (less diversity of candidate states) and as  $|\Delta\mathbf{s}|\rightarrow \infty$ it is
moving toward \textit{exploration} of the optimization landscape
(more diversity of candidate states). This shows importance of using $\Delta\mathbf{s}_{T}$ as an added \textit{early state convergence} metric, since even-though setting $\Delta\mathbf{s}_{T}=200$ has achieved a smaller number of parameters at approximately epoch 160, it decreases the chance of fine-tuning in the remaining
epochs, resulting in an increase in the validation energy
as plotted in Figure~\ref{fig:threshold_earlystate_resenet18_flowers}(e). Since we are minimizing the \textit{energy loss}, it is obvious that $\Delta\mathbf{s}\leq 0$ is guaranteed in (\ref{eq:stateconvergence}). The interesting observation is the pre-mature convergence of $\Delta\mathbf{s}_{T}=50$ and over-fitting of $\Delta\mathbf{s}_{T}=200$ in terms of the best energy and
validation energy in Figures \ref{fig:threshold_earlystate_resenet18_flowers}(d) and \ref{fig:threshold_earlystate_resenet18_flowers}(e). The results in Table~\ref{T:earlystop} suggest $\Delta\mathbf{s}_{T}=100$ is a good
threshold, since it has the highest Top-1 accuracy and also fairly splits half of the training budget for
exploring and the other half for fine-tuning. This threshold is used for the rest of the experiments.

\subsubsection{Number of Candidate State Vectors}
A large number of candidate state vectors (i.e. the population size) $S$ increases the exploration of the optimization landscape and meanwhile the stagnation risk. On the other hand, a small population (${S\leq 8}$,~\cite{salehinejad2017micro}) size encourages exploitation and fine-tuning of optimization. Different methods have been proposed to control this parameter. It has been shown in~\cite{salehinejad2017micro} that it is possible to maintain a high exploration capability while using a small population size by using a vectorized-random mutation factor (VRMF) trick. To analyze $S$, we trained ResNet-18 using \textit{EDropout} for $S\in\{8,16,32,64,128,256\}$ and the Flowers dataset. Table~\ref{T:popsize_resnet18_flowers_EPruning} shows that the small population size $S=8$ and diversified with VRMF achieves the best Top-1 score. However, larger $S$ results in more pruning rate at a lower loss value.  Figure~\ref{fig:pop_EPruning_resenet18_flowers} shows the rate of kept parameter and the best energy of the model during training epochs on the validation dataset, where $\Delta\mathbf{s}_{T}=100$. The plots show that smaller $S$ converges faster to a lower energy while larger $S$ has slower progress and is more prone to stagnation. Since the results show very competitive performance between various values of $S$, we use $S=8$ for the rest of the experiments.

\subsubsection{States Initialization}
The probability $P$ in ${s_{i,d}^{(0)}\sim Bernoulli(P)}$ governs the number of dropped/pruned trainable parameters for each state vector $\mathbf{s}_{i}$ at the initialization stage. If a specific pruning rate is desired, it is possible to define constraints in the evolution phase. Figure~\ref{fig:pinit_EPruning_resenet18_flowers} shows convergence of ResNet-18 with \textit{EDropout} using $P\in\{0.2,0.4,0.6,0.8,1\}$ on the Flowers validation dataset and the corresponding classification results on the test dataset are in Table~\ref{T:init_resnet18_flowers_EPruning}. The results show that at the beginning of training there is diversity in the number of kept parameters but the plots converge to a number in the range of $[45\%,48\%]$. However, this affects the \textit{best energy}, where the model with smaller $P$ converges to a lower energy. Since the results show a little sensitivity of the models to $P$ and convergence to approximately $50\%$ pruning rate, $P$ is set to $0.5$.

\subsection{Classification Performance Analysis}
An original model contains the entire possible trainable parameters and has a larger learning capacity than the corresponding pruned model. \textit{EDropout} is discussed in pruned (P) and full (F) modes. The pruned mode refers to performing inference with the pruned network and the full mode refers to training the original model with EDropout and using the entire trainable parameters for inference. We use the pruning rate chosen by EDropout as the pruning rate for \textit{Deep Compression}~\cite{han2015deep}.

\subsubsection{Kuzushiji Dataset}

Kuzushiji dataset is a fairly easy dataset and performance results of ResNets are very close to each other as shown in Table~\ref{T:results_EPruning}(a). For example, the Top-1 accuracy of ResNet-18 and ResNet-101 is $98.23\%$ and $98.26\%$, respectively, which means overparameterization of ResNets from $11.1$M to $42.5$M only $0.03\%$ increases the Top-1 accuracy. The Top-1 accuracy of ResNet-18+EDropout(P) and ResNet-101+EDropout(P) is $97.78\%$ and $98.05\%$, respectively, where the performance has increased $0.27\%$ and the pruning rate has also increased from $48.51\%$ to $54\%$, respectively. Top-1 accuracy of ResNet-50 is $97.70\%$ and ResNet-50+ThiNet, which has a pruning rate of $55.96\%$, achieves about $2\%$ lower accuracy. ResNet-50+ThiNet has a slightly larger loss value (i.e. $0.2278$) than the ResNet-50+DC (i.e. $0.2142$), but its Top-1 accuracy is $1.38\%$ higher. ResNet-50+EDropout(P) has the closest performance to the original model among the pruning models at a pruning rate of $54.43\%$ and a Top-1 accuracy of $97.63\%$, which is $0.07\%$ lower than ResNet-50.

ChannelNet-v2 has the best Top-1 performance in Table~\ref{T:results_EPruning_alexnet}(a) with $1.8$M parameters. However, its pruning rate is $0\%$ since it is not a pruning method. AlexNet has $57$M parameters and a Top-1 accuracy of $97.77\%$. SqueezeNet is a compact version of AlexNet and has achieved $97.19\%$ Top-1 accuracy with $0.72$M number of parameters. EDropout has achieved a pruning rate of $21.43\%$ for AlexNet, which is inferior to the rate it has achieved in pruning ResNets. AlexNet+EDropout(P) has a Top-1 accuracy of $96.57\%$ which is $1.2\%$ lower than accuracy of AlexNet. EDropout can also be applied to SqueezeNet, where SqueezeNet+EDropout(P) has a Top-1 accuracy of $96.35\%$, which is $0.84\%$ lower than the performance of SqueezeNet. 

\begin{figure*}[!t]
\centering
\captionsetup{font=footnotesize}
\begin{subfigure}[t]{0.24\textwidth}
\centering
\captionsetup{font=footnotesize}
\includegraphics[width=1\textwidth]{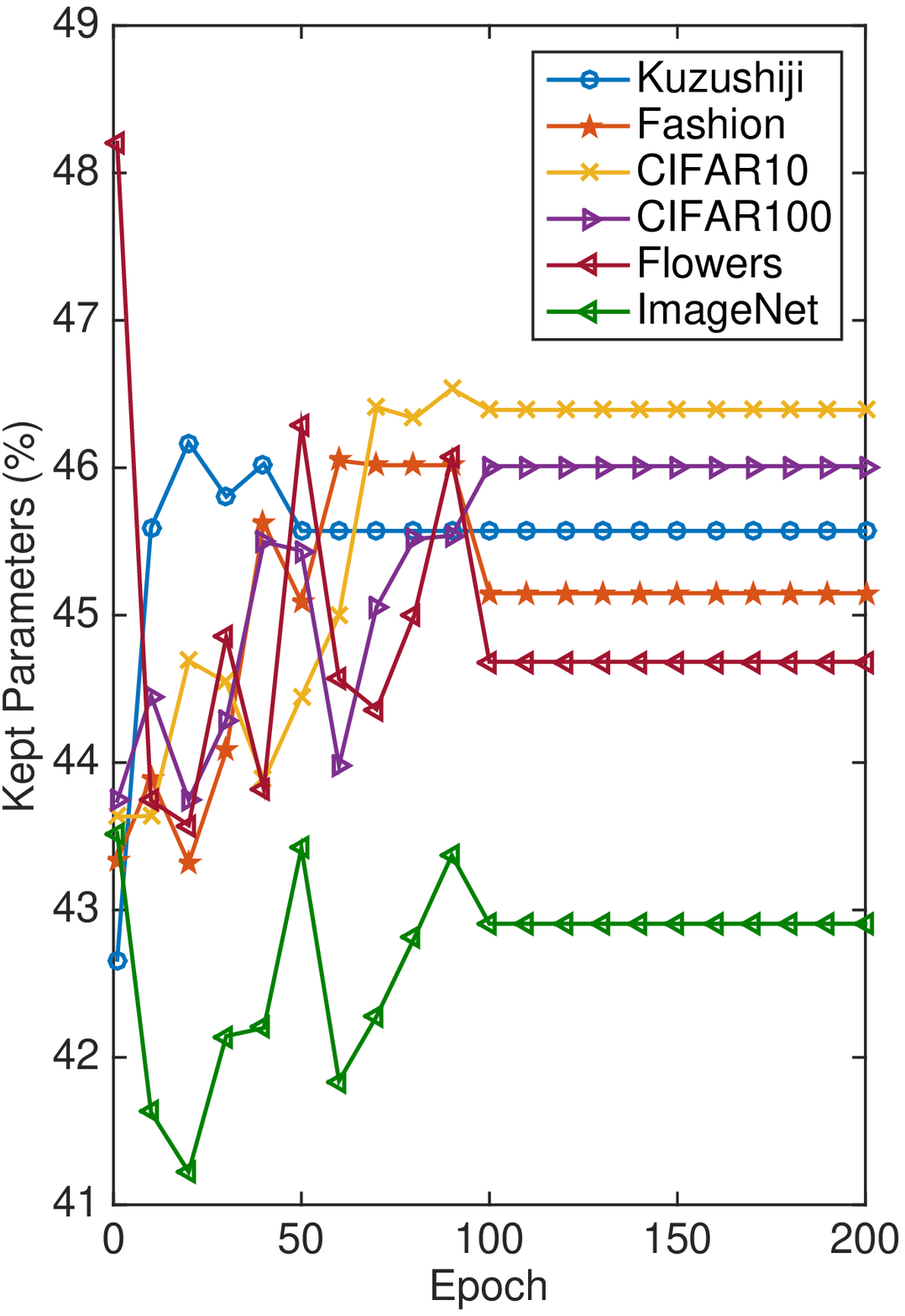}  
\caption{Kept parameters}
\end{subfigure}%
~        
\begin{subfigure}[t]{0.24\textwidth}
\captionsetup{font=footnotesize}
\centering
\includegraphics[width=1\textwidth]{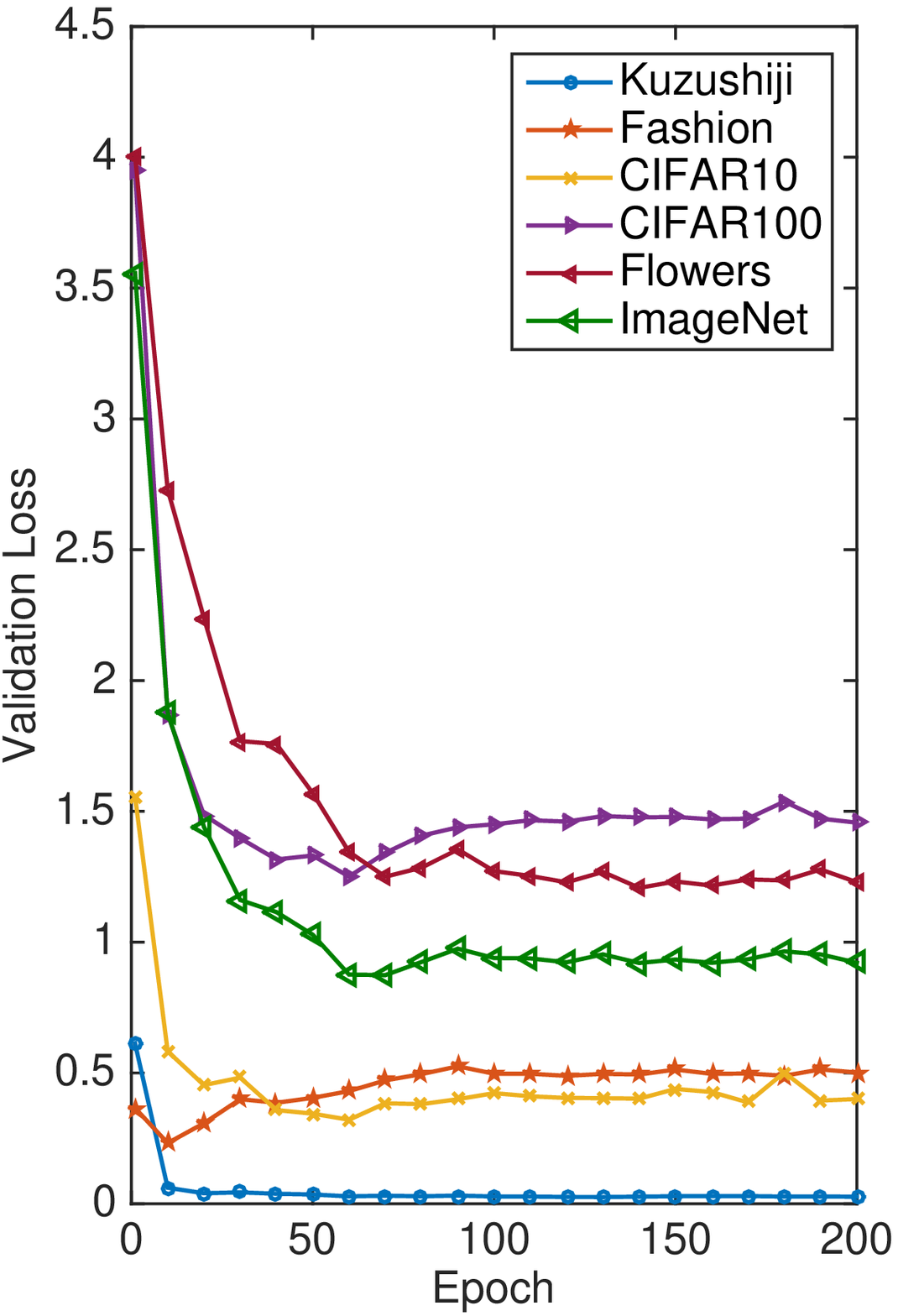}
\caption{Validation loss}
\end{subfigure}%
~    
\begin{subfigure}[t]{0.24\textwidth}
\captionsetup{font=footnotesize}
\centering
\includegraphics[width=1\textwidth]{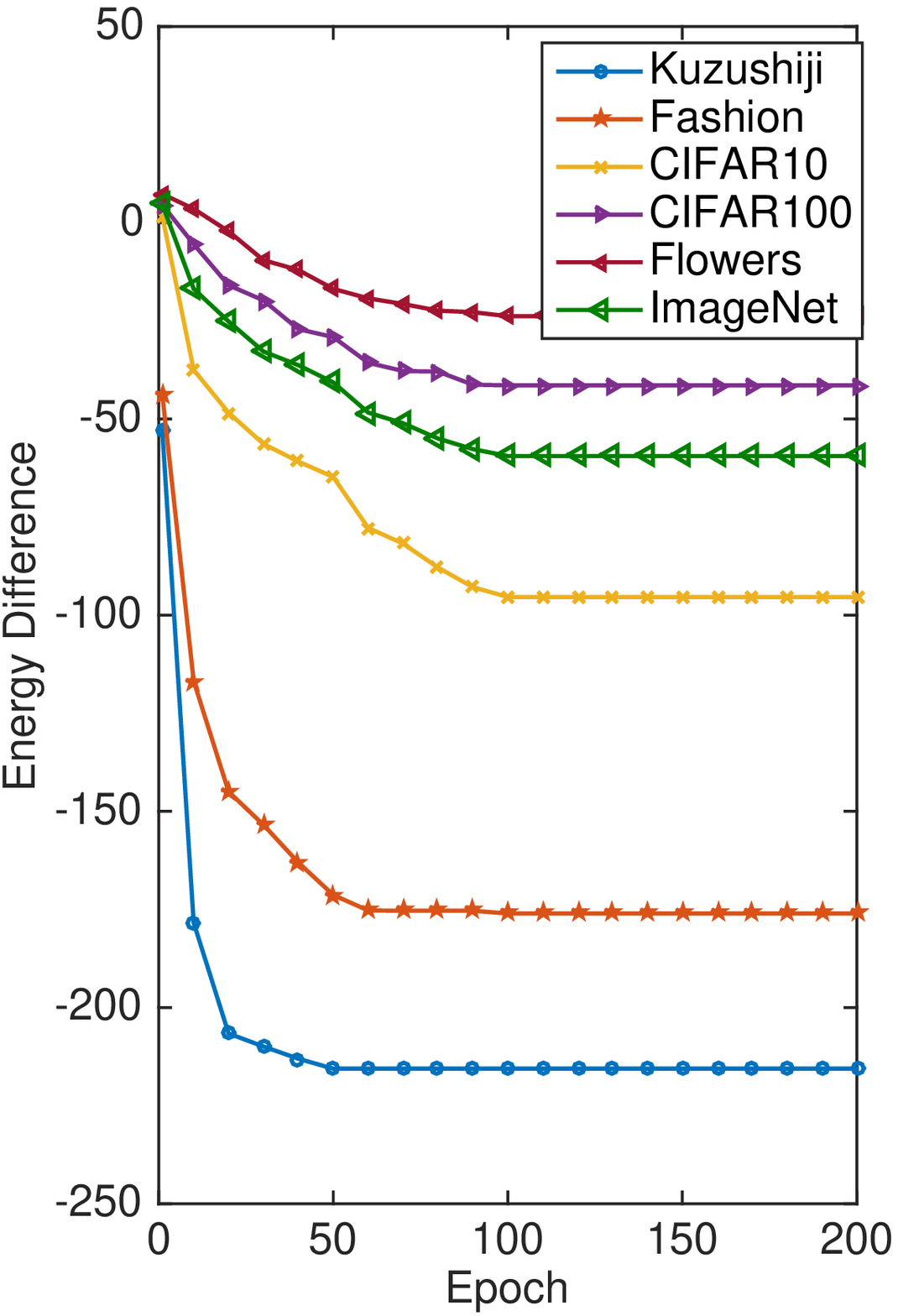}  
\caption{Energy loss difference}
\end{subfigure}%
~
\begin{subfigure}[t]{0.24\textwidth}
\captionsetup{font=footnotesize}
\centering
\includegraphics[width=1\textwidth]{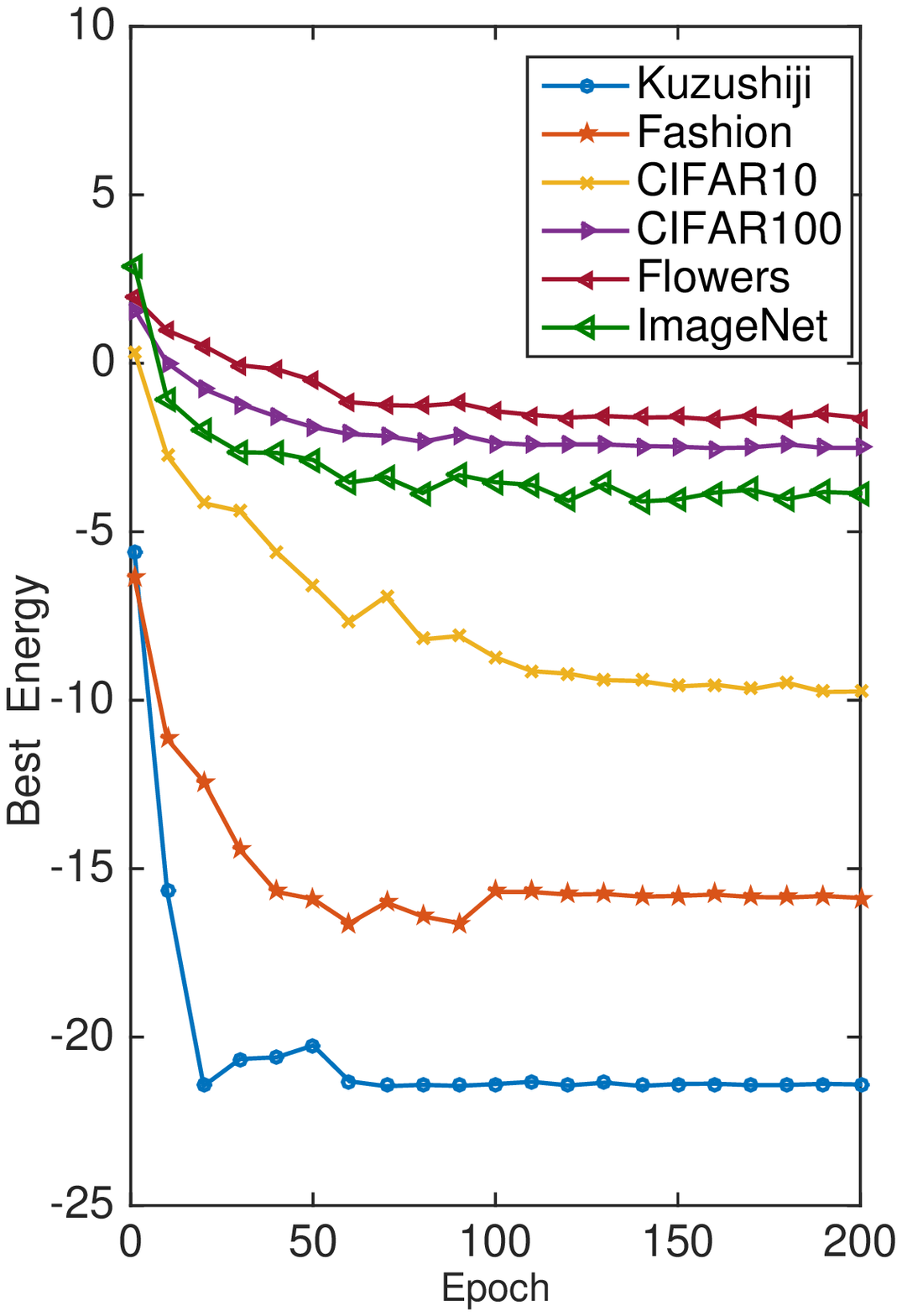}
\caption{Best energy loss}
\end{subfigure}   
\caption{Analysis of \textit{EDropout} with \textbf{ResNet-50} on the validation dataset of Kuzushiji, Fashion, CIFAR-10, CIFAR-100, Flowers, and ImageNet datasets with $S=8$, initialization probability of $P=0.5$, and $\Delta\mathbf{s}_{T}=100$.}
\label{fig:tdatasets_resenet50}
\vspace{-3mm}
\end{figure*}

\subsubsection{Fashion Dataset}

This dataset is more challenging than the Kuzushiji dataset. ResNet-18 has a Top-1 classification performance of $93.87\%$ on the Fashion dataset which is about $5\%$ lower than the Kuzushiji dataset, Table~\ref{T:results_EPruning}(b). ResNet-50+ThiNet has a higher Top-1 accuracy than \textit{Deep Compression} in pruning ResNet-50 with a smaller number of parameters (i.e. $10.3$M). EDropout can prune ResNet-50 at a rate of $54.86\%$ with a Top-1 accuracy of $92.97\%$, which has the best pruning performance for ResNet-50. 

Table~\ref{T:results_EPruning_alexnet}(b) shows that ChannelNet-v2 with $1.7$M number of parameters has a Top-1 accuracy of $93.41\%$. However, SqueezeNet with $0.72$M parameters has a Top-1 accuracy of $92.64\%$ which is only $0.77\%$ lower than ChannelNet-v2. EDropout can further reduce the number of parameters in SqueezeNet to $0.38$M while keeping the performance very close (about $0.3\%$ lower) to the original SqueezeNet.

\subsubsection{CIFAR-10 Dataset}
 Table~\ref{T:results_EPruning}(c) shows that  ResNet-50+ThiNet has a Top-1 accuracy of $79.83\%$ with a pruning rate of $56.18\%$ and ResNet-50 has a Top-1 accuracy of $92.21\%$  with $23.5$M parameters. ResNet-50+DC has approximately $25\%$ lower Top-1 accuracy than the original model. ResNet-50+EDropout(P) has the best performance with a Top-1 accuracy of $85.30\%$ and $10.9$M parameters. 

Table~\ref{T:results_EPruning_alexnet}(c) shows that AlexNet has a Top-1 accuracy of $84.32\%$ with $57.4$M parameters. EDropout applied on the AlexNet model could prune $22.64\%$ of the parameters but the accuracy drops to $74.66\%$. SqueezeNet, which has $0.73$M parameters, has an accuracy of $81.49\%$ where EDropout can prune $47.65\%$ of its parameters with $4.64\%$ drop of Top-1 performance. The results show that EDropout has better performance than the other methods in pruning original models with a slight drop of performance. Particularly, with respect to the Top-3 and Top-5 metrics.

\subsubsection{CIFAR-100 Dataset}
CIFAR-100 has $100$ data classes which increases its classification complexity. ResNet-101 with $42.6$M parameters can achieve $71.19\%$ Top-1 accuracy as shown in Table~\ref{T:results_EPruning}(d). Similar to the previous experiments, EDropout has a better performance than the other methods with a Top-1 classification performance of $61.92\%$ and $18.6$M number of parameters, which has about $9.27\%$ lower Top-1 performance than the ResNet-101 at full capacity. 

Table~\ref{T:results_EPruning_alexnet}(d) shows that AlexNet+EDropout(P) has a Top-1 performance of $56.59\%$ with $41.2$M parameters, which has $3.53\%$ lower Top-1 accuracy than AlexNet with $57.4$M parameters. EDropout can also decrease the number of parameters in SqueezeNet from $0.77$M to $0.43$M with $3.83\%$ drop of Top-1 accuracy. 

\subsubsection{Flowers Dataset}
The Flowers dataset has $102$ number of classes and it is highly limited and imbalanced. Table~\ref{T:results_EPruning}(e) shows that ResNet-101+EDropout(F) has the best Top-1 accuracy for ResNet-101 with $3.26\%$ drop of performance. The pruned version of EDropout has an accuracy of $58.17\%$ with $19.2$M number of parameters, which is $44.85\%$ of the ResNet-101 parameters.

Table~\ref{T:results_EPruning_alexnet}(e) shows that AlexNet has $11\%$ higher Top-1 accuracy than SqueezeNet on the Flowers dataset, where AlexNet has $57.4$M parameters and SqueezeNet has $0.77$M parameters. SqueezeNet+EDropout(P) reduces the number of parameters in SqueezeNet to $0.57$M with almost $2.22\%$ of drop in Top-1 accuracy.

\subsubsection{ImageNet Dataset}
Table~\ref{T:results_EPruning}(f) shows that EDropout has a relatively higher Top-1 accuracy than the other pruning models with respect to the pruning rate. ResNet-50+ThiNet has an accuracy of $72.11\%$ with $12.4$M parameters and ResNet-50+DC has an accuracy of $67.02\%$ with $10.2$M parameters, which are $3.16\%$ and $8.25\%$ lower than the Top-1 performance of ResNet-50, respectively. This is while ResNet-50+EDropout(P) has $1.55\%$ lower performance than ResNet-50 with $10.2$M parameters.

ChannelNet-v2 in Table~\ref{T:results_EPruning_alexnet}(f) has a Top-1 accuracy of $69.5\%$  with $2.7$M parameters. SqueezeNet has an accuracy of $57.31\%$ with $1.3$M parameters and SqueezeNet+EDropout(P) achieves $0.7\%$ lower accuracy by pruning $39.68\%$ of the parameters. AlexNet has a Top-1 accuracy of $58.26\%$ with $60$M parameters and AlexNet+EDropout(P) decreases the number of parameters in AlexNet to $49.2$M at the cost of $2.23\%$ lower Top-1 performance.

\subsection{Convergence Analysis}
Figure~\ref{fig:tdatasets_resenet50} shows convergence plots of the validation datasets for ResNet-50 with \textit{EDropout}. Figure~\ref{fig:tdatasets_resenet50}(a) shows that after epoch $100$ the \textit{early state convergence} is applied. EDropout+ResNet-50  converges to a range approximately between $43\%$ and $47\%$ of the total number of parameters in the original model. The CIFAR-100, Flowers, and ImageNet datasets have higher validation loss than the other datasets in Figure~\ref{fig:tdatasets_resenet50}(b). EDropout+ResNet-50 converges faster for simple datasets such as Fashion comparing with the more challenging ones. 
The other observation is that complex datasets, in terms of number of classes, number of training samples, number of image channels, and input image size, have higher \textit{best energy} values, which reflects more number of parameters are required to train a model to perform classification in these datasets. As an example, the Flowers dataset in Figure~\ref{fig:tdatasets_resenet50}(d) has the highest \textit{best energy} values while the Kuzushiji dataset has the lowest energy values. Similar trend is observable in the \textit{energy loss difference} (i.e. difference between the \textit{best energy} of the population and the \textit{average energy} of the population) plots in Figure~\ref{fig:tdatasets_resenet50}(c).

\begin{table}[!tbp]
\centering
\captionsetup{font=footnotesize}
\caption{Training time (in seconds/iteration) for the population size $S\in\{8,32\}$ and a batch size of 128 using CIFAR-10.\\ 
\centerline{\fcolorbox{black}[HTML]{9AFF99}{\rule{0pt}{6pt}\rule{6pt}{0pt}}\;E-3\quad 
\fcolorbox{black}[HTML]{32CB00}{\rule{0pt}{6pt}\rule{6pt}{0pt}}\;E-2\quad
\fcolorbox{black}[HTML]{009901}{\rule{0pt}{6pt}\rule{6pt}{0pt}}\;E-1} 
}
\begin{adjustbox}{width=0.44\textwidth}
\begin{tabular}{|c|c|c|c|c|c|}
\hline
Model &
  Original &
  \begin{tabular}[c]{@{}c@{}}EDropout\\ S=8\end{tabular} &
  \begin{tabular}[c]{@{}c@{}}EDropout\\ S=32\end{tabular} &
  \begin{tabular}[c]{@{}c@{}}EDropout\\ Parallel\\ S=8\end{tabular} &
  \begin{tabular}[c]{@{}c@{}}EDropout\\ Parallel\\ S=32\end{tabular} \\ \hline
AlexNet &
  \cellcolor[HTML]{9AFF99}4.14E-3 &
  \cellcolor[HTML]{32CB00}1.26E-2 &
  \cellcolor[HTML]{32CB00}5.06E-2&
  \cellcolor[HTML]{9AFF99}8.01E-3 &
  \cellcolor[HTML]{32CB00}2.67E-2 \\ \hline
ResNet-18 &
  \cellcolor[HTML]{9AFF99}7.11E-3 &
  \cellcolor[HTML]{32CB00}2.93E-2 &
  \cellcolor[HTML]{009901}1.19E-1 &
  \cellcolor[HTML]{32CB00}2.15E-2 &
  \cellcolor[HTML]{32CB00}6.33E-2 \\ \hline
ResNet-34 &
  \cellcolor[HTML]{32CB00}1.13E-2 &
  \cellcolor[HTML]{32CB00}5.03E-2 &
  \cellcolor[HTML]{009901}1.87E-1 &
  \cellcolor[HTML]{32CB00}3.62E-2 &
  \cellcolor[HTML]{32CB00}8.28E-2 \\ \hline
ResNet-50 &
  \cellcolor[HTML]{32CB00}1.63E-2 &
  \cellcolor[HTML]{32CB00}7.46E-2 &
  \cellcolor[HTML]{009901}2.90E-1 &
  \cellcolor[HTML]{32CB00}5.45E-2 &
  \cellcolor[HTML]{009901}1.42E-1 \\ \hline
ResNet-101 &
  \cellcolor[HTML]{32CB00}5.20E-2 &
  \cellcolor[HTML]{009901}1.62E-1 &
  \cellcolor[HTML]{009901}5.46E-1 &
  \cellcolor[HTML]{009901}1.14E-1 &
  \cellcolor[HTML]{009901}2.74E-1 \\ \hline

\end{tabular}
\end{adjustbox}
\label{T:time_EPruning}
\vspace{-4mm}
\end{table}

\subsection{Computational Complexity Analysis}
The main drawback of EDropout is its computational complexity as a result of using a population of state vectors in the optimization phase to prune a network. 

The major operations in training a neural network are feedforward (inference) and backpropagation. In order to provide a theoretical framework, let us define the feedforward time per iteration as $T_{inf}$ and the backpropagation time per iteration as $T_{bp}$. We also assume ${T_{inf}(EDropout)\approx T_{inf}(Net)}$ and ${T_{bp}(EDropout)\approx T_{bp}(Net)}$ for simplicity, where $Net$ refers to the original network. Hence, we can define the total training time of an original network as
\begin{equation}
T(Net)=N_{epoch}N_{batch}(T_{inf}+T_{bp}).
\end{equation}
 EDropout has another major operation which is the \textit{evolution} of the population of \textit{candidate state vectors}. Hence, the training time of EDropout applied on an original network is
\begin{equation}
T(EDropout)=N_{epoch}N_{batch}(S\cdot T_{inf}+T_{bp}+T_{evn}),
\end{equation}
where $T_{evn}$ is the \textit{evolution} time per training iteration.

\begin{figure}[!tbp]
\centering
\captionsetup{font=footnotesize}
\begin{subfigure}[t]{0.48\textwidth}
\centering
\captionsetup{font=footnotesize}
\includegraphics[width=1\textwidth]{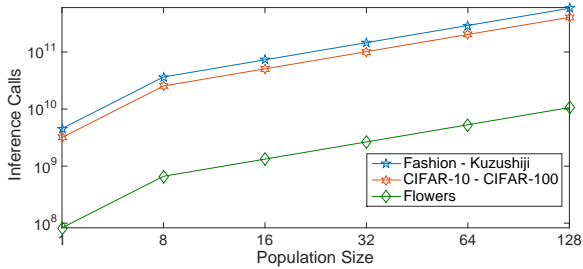}  
\caption{Upper boundary of number of inference calls for different population sizes $S$ and different number of training samples (the datasets in one line of legend have similar number of training samples). The Inference Calls axis is log scaled.}
\label{fig:EPruning_inferencecalls}
\end{subfigure}%
\vspace{3mm}

\begin{subfigure}[!t]{0.48\textwidth}
\captionsetup{font=footnotesize}
\centering
\includegraphics[width=1\textwidth]{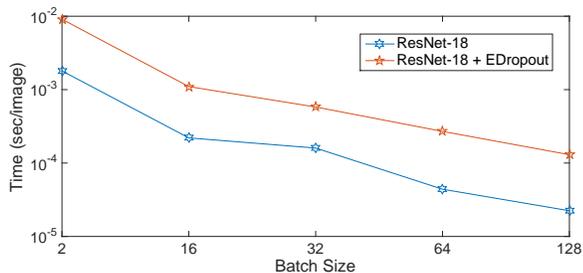}  
\caption{Inference time of ResNet-18 for different data batch sizes in seconds/image. The Time axis is log scaled.}
\label{fig:EPruning_time}
\end{subfigure}%
\caption{Computational complexity of EDropout with respect to the number of inference calls and inference time per image.}
\label{fig:EPruning_time_complexity}
\vspace{-6mm}
\end{figure}

In the worst-case scenario (i.e. without considering the \textit{early state convergence}), for $N_{epoch}$ number of epochs and $N_{batch}$ number of iterations per epoch, the upper bound for the number of inference calls (i.e. performing inference) is $S\cdot N_{epoch}\cdot N_{batch}$, that is $S$ times more than one inference per training iteration, as demonstrated in Figure~\ref{fig:EPruning_time_complexity}(a) for 200 training epochs. By considering the \textit{early state convergence} condition met after $\Delta s_{T}$ training epochs, $T(EDropout)$ becomes
\begin{equation}
\begin{split}
   T(EDropout;\Delta s_{T})&=\underbrace{\Delta s_{T}\cdot N_{batch}(S\cdot T_{inf}+T_{bp}+T_{evn})}_\text{exploration}\\
   &+\underbrace{(N_{epoch}-\Delta s_{T})\cdot N_{batch}\cdot(T_{inf}+T_{bp})}_\text{fine-tuning}\\
   &=\underbrace{N_{epoch}N_{batch}(T_{inf}+T_{bp})}_\textit{T(Net)} \\
   &+\underbrace{N_{batch}\Delta s_{T}\big(T_{evn}+(S-1)T_{inf}\big)}_\text{EDropout overhead},
\end{split}
\end{equation}
where $\Delta s_{T}\leq N_{epoch}$.

Table~\ref{T:time_EPruning} shows the comparison between $T(EDropout)$ and $T(Net)$ per training iteration for a batch size of 128 using the CIFAR-10 dataset. The benefit of the computational overhead of EDropout is the added \textit{diversity} in searching for the sub-network with lower energy. Since the candidate states in the population are independent, one can implement the optimizer in parallel at the state vector level, which shows dramatic reduction of the $T(EDropout)$ in Table~\ref{T:time_EPruning}. 

 It is obvious that increasing the data batch size for each iteration can also decrease the training time. As an example, Figure~\ref{fig:EPruning_time_complexity}(b) shows that increasing the batch size can decrease the training and inference time of ResNet-18. Another approach which reduces the execution time is using the concept of \textit{early state convergence}, where the number of iterations for sub-network exploration can be defined or adaptively set by the model according to~(\ref{eq:stateconvergence}), as discussed in Subsection~\ref{sec:earlystateconvergence}.

\section{Conclusions}
\label{sec:conclusions}
In this paper, we introduce the concept of energy-based dropout and pruning in deep neural networks. We propose a new method for partial training of deep neural networks (DNNs) based on the concept of dropout. An energy model is proposed to compute energy loss of a population of binary candidate pruning state vectors, where each vector represents a sub-network of the DNN. An evolutionary technique searches the population of state vectors, selects the state vector with lowest energy loss in that iteration, and trains the corresponding sub-network using backpropagation. Ultimately, the states can converge to a \textit{best state} and the algorithm continues fine-tuning the corresponding sub-network, which is equivalent to \textit{pruning} of the DNN. The results show that there is a trade-off between network size and
classification accuracy. The proposed method
can reduce (approximately $50\%$ on average) the number of parameters in the inference network while keeping the classification accuracy competitive to the original network.



\section{Acknowledgment}
The authors acknowledge financial support and access to the Digital Annealer (DA) of Fujitsu Laboratories Ltd. and Fujitsu Consulting (Canada) Inc.

\bibliographystyle{IEEEtran}
\bibliography{CTLIEEEtrans}

\begin{IEEEbiography}
[{\includegraphics[width=1\textwidth]{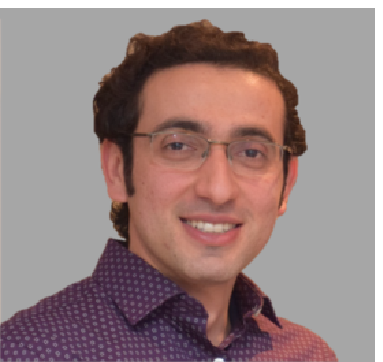}}]{Hojjat Salehinejad} is a PhD candidate at the Edward S. Rogers Sr. Department of Electrical and Computer Engineering, University of Toronto. His research focus is on machine learning and signal processing. He is a member of IEEE Signal Processing Society and has served as a reviewer for various IEEE conferences and journals. 
\end{IEEEbiography}
\vskip 5pt plus -1fil
\begin{IEEEbiography}
[{\includegraphics[width=1\textwidth]{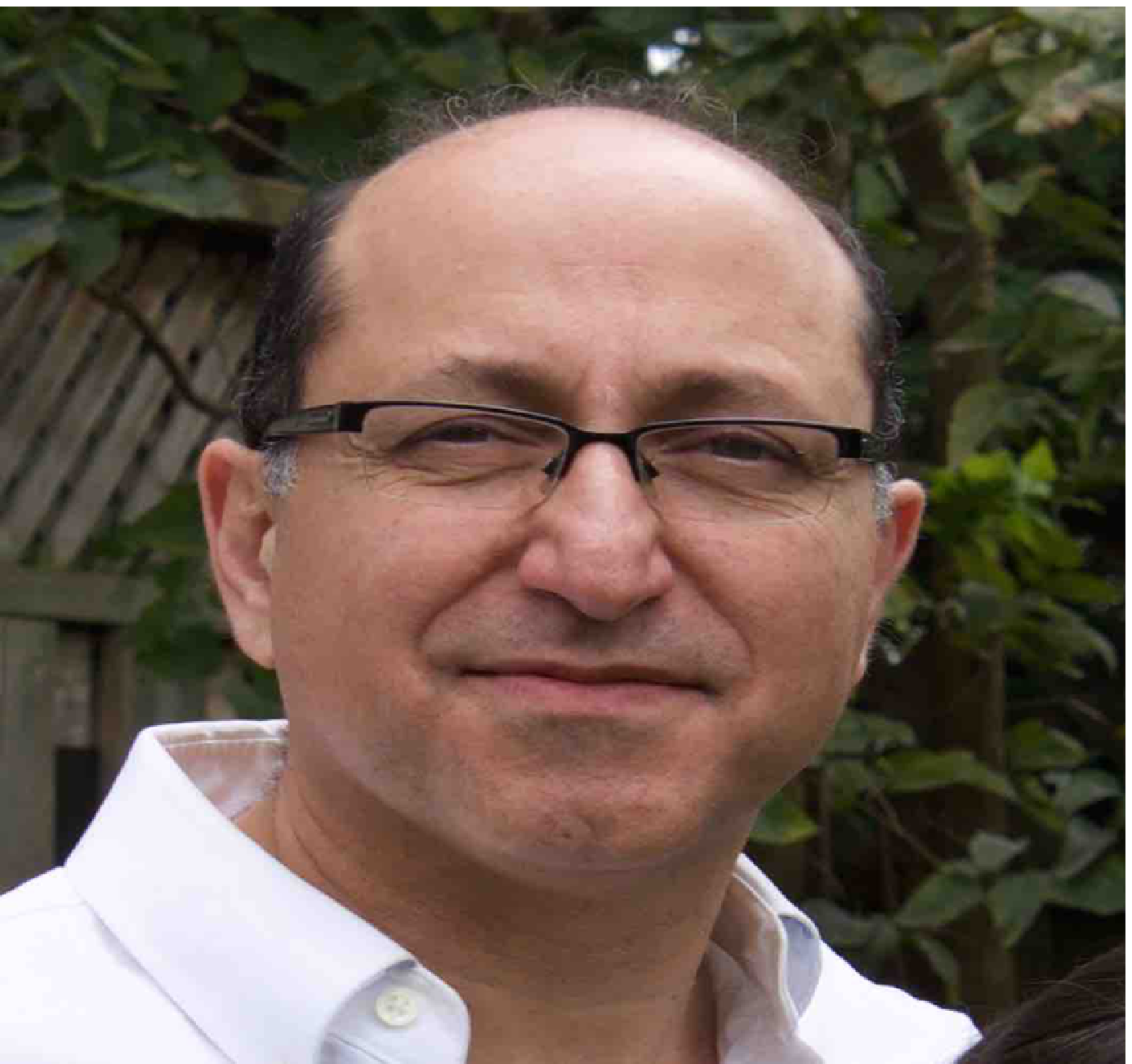}}]{Shahrokh Valaee} is a Professor with the Edward S. Rogers Sr. Department of
Electrical and Computer Engineering, University of Toronto, and the holder of Nortel
Chair of Network Architectures and Services. He is the Founder and the Director of the
Wireless and Internet Research Laboratory (WIRLab) at the University of Toronto.
Professor Valaee was the TPC Co-Chair and the Local Organization Chair of the IEEE
Personal Mobile Indoor Radio Communication (PIMRC) Symposium 2011. He was the
TCP Chair of PIMRC2017, the Track Co-Chair of WCNC 2014, the TPC Co-Chair of
ICT 2015. He has been the guest editor for various journals. He was a Track Co-chair
for PIMRC 2020 and VTC Fall 2020. From December 2010 to December 2012, he was
the Associate Editor of the IEEE Signal Processing Letters. From 2010 to 2015, he
served as an Editor of IEEE Transactions on Wireless Communications. Currently, he is
an Editor of Journal of Computer and System Science. Professor Valaee is a Fellow of
the Engineering Institute of Canada, and a Fellow of IEEE.

\end{IEEEbiography}

\end{document}